\documentclass[10pt,twocolumn,letterpaper]{article}

\usepackage{cvpr}              % To produce the CAMERA-READY version
\usepackage[dvipsnames]{xcolor}

\definecolor{cvprblue}{rgb}{0.21,0.49,0.74}
\usepackage[pagebackref,breaklinks,colorlinks,citecolor=cvprblue]{hyperref}

\usepackage{tabularx}
\usepackage{mathrsfs}
\usepackage{xcolor,colortbl}
\usepackage{bbding}
\usepackage{multirow}

\def\paperID{7967} % *** Enter the Paper ID here
\def\confName{CVPR}
\def\confYear{2024}

\usepackage{xcolor}

\title{DualNeRF: Text-Driven 3D Scene Editing via Dual-Field Representation}

\author{Yuxuan Xiong\\
\and
Yue Shi\\
\and
Yishun Dou\\
\and
Bingbing Ni
}

\begin{document}

\maketitle

\begin{abstract}
Recently, denoising diffusion models have achieved promising results in 2D image generation and editing.
Instruct-NeRF2NeRF (IN2N) introduces the success of diffusion into 3D scene editing through an ``Iterative dataset update" (IDU) strategy.
Though achieving fascinating results, IN2N suffers from problems of blurry backgrounds and trapping in local optima.
% Though achieving fascinating results, IN2N generates edits with blurry backgrounds.
The first problem is caused by IN2N's lack of efficient guidance for background maintenance, while the second stems from the interaction between image editing and NeRF training during IDU.
In this work, we introduce \textbf{DualNeRF} to deal with these problems.
We propose a dual-field representation to preserve features of the original scene and utilize them as additional guidance to the model for background maintenance during IDU.
% We introduce a dual-field representation to stabilize the training process and greatly enhance the visual quality of unedited areas, which usually look blurry for IN2N. 
Moreover, a simulated annealing strategy is embedded into IDU to endow our model with the power of addressing local optima issues.
% A CLIP-based consistency indicator is also used to filter out low-quality edits.
A CLIP-based consistency indicator is used to further improve the editing quality by filtering out low-quality edits.
Extensive experiments demonstrate that our method outperforms previous methods both qualitatively and quantitatively.
\end{abstract}    
\section{Introduction}
\label{sec:intro}

% Recently, neural radiance field (NeRF) \cite{mildenhall2021nerf}, which represents a target scene by a multi-layer perception (MLP) model and its parameters, has gained more and more attention in the field. Due to the strong representation ability and high compactness of MLP, NeRF and its follow-ups \cite{zhang2020nerf++, liu2020neural, barron2021mip} achieves better scene reconstruction quality with a lower memory budget compared to traditional explicit 3D representations. What's more, the differentiable volume rendering algorithm enables NeRF to be trained end-to-end based on multi-view images along with their corresponding camera parameters.

% Neural radiance field (NeRF) \cite{mildenhall2021nerf}, which represents a target object (or scene) by neural networks (NNs) and their parameters, has obtained increasing interest. Thanks to the strong representation ability and high compactness of NNs, NeRF and its follow-ups \cite{zhang2020nerf++, liu2020neural, barron2022mip, muller2022instant} offer enhanced reconstruction quality with a lower memory footprint in comparison to traditional explicit 3D representations. However, the editing of NeRF is a notorious problem, due to the entanglement between different components represented implicitly in the field.

% \yue{sentence1: start with the meaning of 3D editing . sentence2: there have been a series 3D editing projects with the development of 3D representation and generative model(such as diffusion). sentence3: however, problem.}

3D implicit scene editing constitutes a significant yet challenging task in the realm of computer graphics and computational vision, intending to modify an existing 3D scene represented by an implicit field. The advancement of implicit 3D representations has fostered a myriad of 3D editing endeavors, including \cite{yuan2022nerf, garbin2022voltemorph, xu2022deforming, tertikas2023partnerf, peng2022cagenerf, jambon2023nerfshop, liu2021editing, xiang2021neutex, gong2023recolornerf, kuang2023palettenerf, lee2023ice, niemeyer2021giraffe, xu2023discoscene, li2022climatenerf, zhang2021nerfactor, boss2021neural, srinivasan2021nerv, zhang2022arf, wang2023nerf}. Nevertheless, most of them have concentrated on rudimentary modifications, such as geometry or texture editing, which restricts the generalizability and user accessibility of these methods.

% Furthermore, NeRF's utilization of a differentiable volume rendering algorithm facilitates end-to-end training based on multi-view images and their corresponding camera parameters.

% However, the editing of NeRF is a notorious problem. Since NeRF represents the target object implicitly by MLP parameters, different components and attributes of the object tend to entangle with each other, which makes it more difficult to edit the target object compared to the traditional explicit representation cases. Many previous works have proposed solutions for NeRF editing, concentrating on the tasks of geometry deformation \cite{yuan2022nerf, garbin2022voltemorph, xu2022deforming, tertikas2023partnerf, peng2022cagenerf, jambon2023nerfshop}, texture or color editing \cite{liu2021editing, xiang2021neutex, gong2023recolornerf, kuang2023palettenerf, lee2023ice}, scene manipulation \cite{niemeyer2021giraffe, xu2023discoscene}, relighting \cite{li2022climatenerf, zhang2021nerfactor, boss2021neural, srinivasan2021nerv} and stylization \cite{zhang2022arf, wang2023nerf}.

%Recently, as the emergence of vision-language models, such as CLIP \cite{radford2021learning} and noise diffusion model \cite{nichol2021glide, rombach2022high, saharia2022photorealistic, ramesh2022hierarchical}, more and more researchers utilize pre-trained 2D text-image models to edit implicit neural fields based on user-entered text instructions \cite{wang2022clip, wang2023nerf, haque2023instruct, fang2023text, mirzaei2023watch}.
Recent advancements in vision-language models, notably CLIP \cite{radford2021learning} and various noise diffusion models \cite{nichol2021glide, rombach2022high, saharia2022photorealistic, ramesh2022hierarchical}, have prompted an increase in the use of pre-trained 2D text-image models for editing implicit neural fields via text instructions \cite{wang2022clip, wang2023nerf, haque2023instruct, fang2023text, mirzaei2023watch}. 
The Instruct-NeRF2NeRF (IN2N) framework \cite{haque2023instruct}, utilizing the "Iterative Dataset Update" (IDU) strategy, represents a significant development in this area.
% Instruct-NeRF2NeRF (IN2N) \cite{haque2023instruct} proposes a neural field editing framework based on the ``Iterative Dataset Update" (IDU) strategy.
IDU leverages a 2D text-based editing model, Instructpix2pix (IP2P) \cite{brooks2023instructpix2pix}, to update the training dataset and finetunes the neural fields alternatively. In this way, both the dataset and model are updated to align to a user-provided text prompt. Despite the fascinating editing results, IN2N reveals several limitations.

\begin{figure}
    \centering

    \rotatebox{90}{~~ Blurry Background}
    \begin{subfigure}{0.31\linewidth}
        \begin{minipage}[t]{1.0\linewidth}
            \centering
            \includegraphics[width=1.0\linewidth]{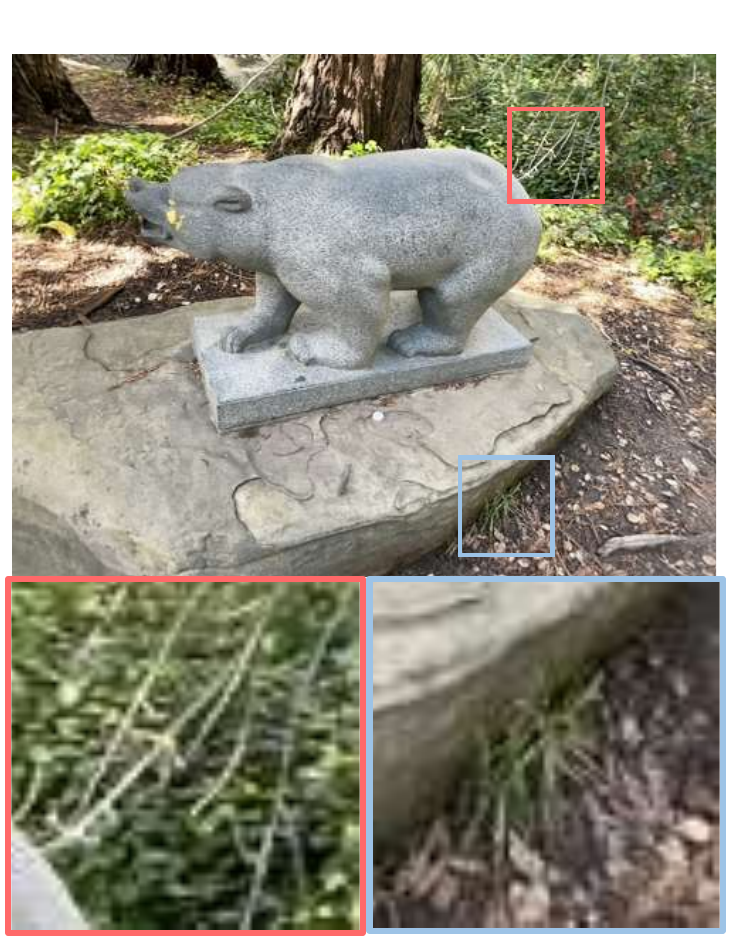}
            \caption{Original Scene}
            \label{fig: 1-original}
        \end{minipage}
    \end{subfigure}
    \hfill
    \begin{subfigure}{0.31\linewidth}
        \begin{minipage}[t]{1.0\linewidth}
            \centering
            \includegraphics[width=1.0\linewidth]{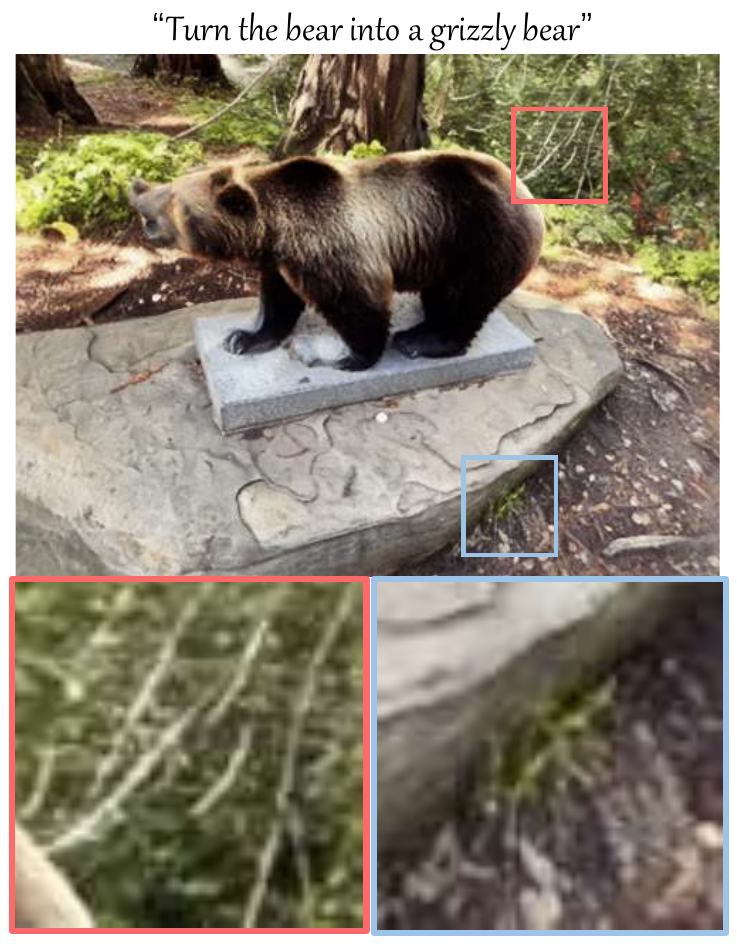}
            \caption{IN2N}
            \label{fig: 1-IN2N}
        \end{minipage}
    \end{subfigure}
    \hfill
    \begin{subfigure}{0.31\linewidth}
        \begin{minipage}[t]{1.0\linewidth}
            \centering
            \includegraphics[width=1.0\linewidth]{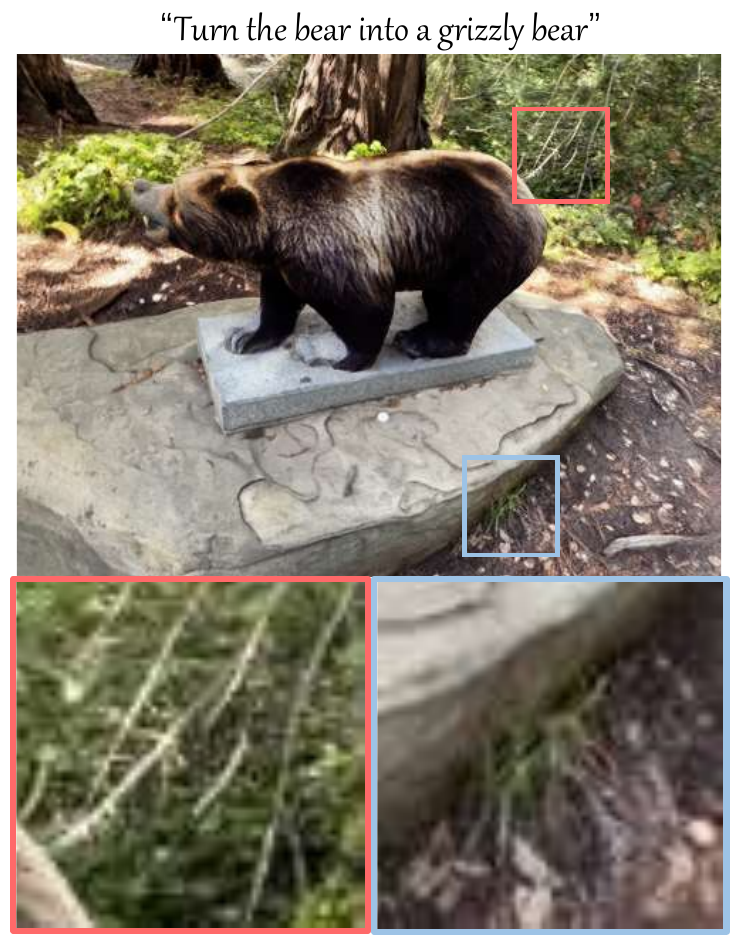}
            \caption{DualNeRF}
            \label{fig: 1-DualNeRF}
        \end{minipage}
    \end{subfigure}

    \rotatebox{90}{~Local Optima}
    \begin{subfigure}{0.95\linewidth}
        \begin{minipage}[t]{1.0\linewidth}
            \centering
            \includegraphics[width=1.0\linewidth]{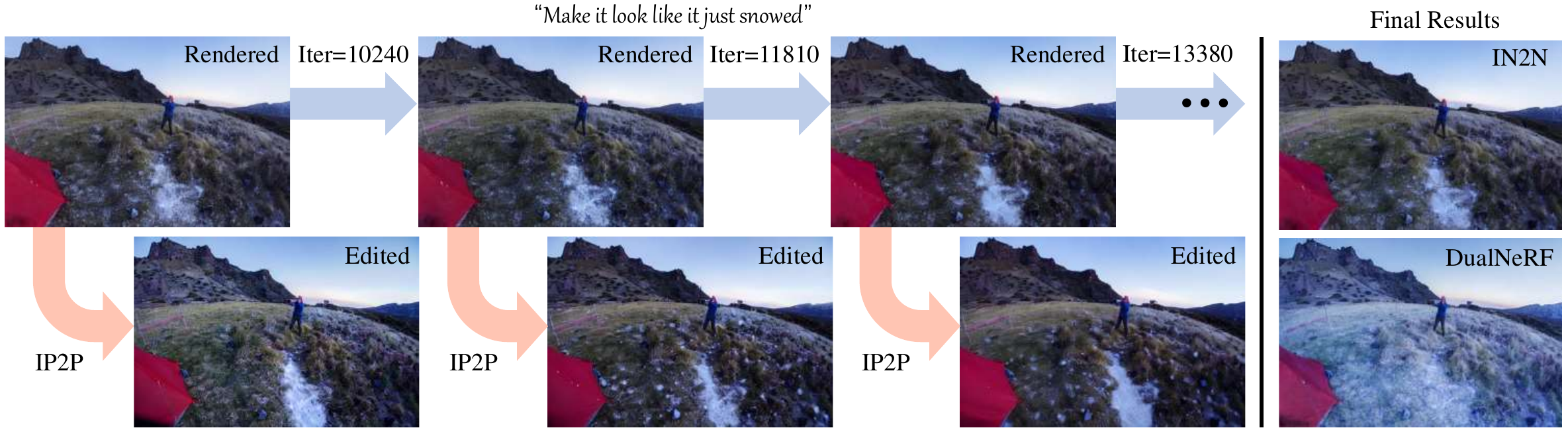}
            \caption{An example of IN2N's local optima issues}
            \label{fig: 4-local_optima}
        \end{minipage}
    \end{subfigure}
  
    \caption{\textbf{Limitations of IN2N \cite{haque2023instruct}.} There are two main limitations exposed by IN2N: (1) blurry background and (2) being prone to the local optima. The first row shows a comparison of the background performance among the rendering results of the original scene, IN2N, and ours. IN2N generates the most blurry background. The second row shows an example of IN2N's local optima issues which manifests as incomplete edits to the original scene. In comparison, DualNeRF outputs satisfactory results.}
    \label{fig: IN2N_drawbacks}
\end{figure}

% Despite the fantastic final editing results, IN2N models usually fall into a chaotic intermediate state with obvious artifacts during training. These artifacts are caused by the IDU strategy which alternately updates the train data and inevitably brings inconsistency into the training dataset. IN2N models with these artifacts generate degenerate rendering results, which further mislead IP2P to generate low-quality editing results and contaminate the training dataset with more inconsistency. An example of this series of events is shown in s figure \ref{fig: artifacts}.

% Despite the fascinating editing results, we show by experiments that IN2N exposes drawbacks including blurry backgrounds and easily trapped in local optima, as shown in Fig. \ref{fig: IN2N_drawbacks}. These drawbacks severely hurt the final editing quality of IN2N. In this work, we propose a new text-driven 3D scene editing method called \textbf{DualNeRF} to deal with these problems.

Firstly, IN2N generates edited scenes with blurry backgrounds, as shown in Fig. \ref{fig: 1-IN2N}. Essentially, IDU is a training process that optimizes both the dataset and the model with text prompt $y$ as the only guidance to control the optimization direction. This weak guidance provides no guarantee of preserving the original background. IP2P edits with distorted backgrounds provide wrong training signals to the NeRF model, jittering the original texture and resulting in blurred backgrounds after training. The single-field architecture used in IN2N cannot deal with this problem, since it is unable to preserve any initial feature after long-term optimizations.

% Secondly, IN2N is prone to be trapped in a local optimal solution. As shown in Fig. \ref{fig: 4-local_optima}, the incompletely edited rendering results (shown in the top row) mislead IP2P to generate edits with similar appearances (shown in the bottom row). This in turn provides wrong training signals to the model and makes the situation even worse. After a long period of training, the incomplete editing issue becomes ineffaceable. This marks that the model traps in local optima. In other words, IDU tends to preserve some sub-optimal edits during training due to the mutual reinforcement between IP2P edits and model training.

Secondly, the IN2N framework exhibits a susceptibility to becoming ensnared in local optima. As illustrated in Fig. \ref{fig: 4-local_optima}, partially edited renderings (displayed in the top row) mislead IP2P to produce edits with similar appearances (displayed in the bottom row). This in turn provides wrong training signals to the model and makes the situation even worse. Over prolonged training duration, the incomplete editing issue becomes ineffaceable, signifying the model's entrapment in local optima. In essence, IDU tends to preserve sub-optimal edits during training due to the mutual reinforcement between IP2P edits and model training.

% First of all, we propose a dual-field representation called \textbf{DualNeRF} to stabilize the training process and therefore preserve more background details. Specifically, DualNeRF contains two neural networks. One is set as the major field, whose parameters are trained to faithfully reconstruct the original scene and frozen during editing. Another is designed as the editing field, which is gradually enabled and trained during editing. Output features from the two fields fused by activation functions output the final result. DualNeRF performs flexible editing by the editing field while preserving the information of the original scene by the major field to stabilize the quality of the unedited area during training.

We show by experiments that these limitations severely hurt the final editing quality of IN2N. In this work, we propose a novel text-driven 3D scene editing method called \textbf{DualNeRF} to address these problems.

% First of all, we propose a dual-field representation to deal with the blurry background problem. Specifically, DualNeRF consists of two fields. One of them is trained to reconstruct the original scene and frozen during editing, while the other one is optimized for edits. The frozen field preserves authentic features of the original scene. These features serve as additional guidance during IDU, avoiding the NeRF model drifting away from the original scene caused by the background distortion in IP2P edits after long-term optimization.

% First of all, we introduce new guidance signals to the model during IDU to help preserve texture in background areas. Intuitively, the initial field before editing contains abundant features of the original scene, which can serve as perfect guidance for background maintenance. However, these features drift away from the initial stage during the long-term optimization of IDU. To preserve these features during training, we propose a dual-field representation. One field in the model serves as the guidance field which is frozen during IDU and provides features of the original scene. The other field serves as an editing field which is trained to perform edits during IDU. Features provided by the guidance field prevent the model from drifting away from the original scene, countering background distortion brought by IP2P edits.

First of all, we introduce additional guidance signals to the model during IDU to maintain textures in the background areas.
% Recognizing that abundant features of the original scene in the initial field offer excellent background preservation guidance, we propose a dual-field representation to combat feature drift during the long-term optimization of IDU.
Intuitively, the initial field before editing contains abundant features of the original scene, which can serve as perfect guidance for background maintenance. However, these features drift away from the initial stage during the long-term optimization of IDU. To preserve these features during training, we propose a novel dual-field representation.
This representation comprises a static field, preserving the original scene's features for guidance, and a dynamic field, trained for performing edits. Features provided by the static field help stabilize the model, mitigating background distortions often induced by IP2P edits.

Moreover, a simulated annealing (SA) strategy \cite{kirkpatrick1983optimization} is incorporated into IDU to address local optima issues. SA is a well-known algorithm used to solve local optima through random acceptance of sub-optimal updates. Inspired by this idea, instead of editing the renderings from the latest model, we randomly send some ``outdated" inputs to IP2P for editing. These outdated inputs are derived from ``half-edited" models by decreasing the intensity of the dynamic field, which will be introduced in details in Sec. \ref{sec: Simulated Annealing Strategy}.
% This simulated annealing strategy helps our model jump out of local optima efficiently.
This adaptation of the SA strategy significantly enhances our model's ability to overcome local optima.

A CLIP-based consistency indicator is also used to measure the reliability of each editing result of IP2P. Editing results with higher reliability are controlled to exert stronger impacts on the neural field, and vice versa. In this way, high-quality editing results with fewer artifacts will be ``reserved", while low-quality results which extremely deviate from the original image will be ``filtered out".

In summary, the contributions of our work include:
\begin{itemize}
    \item We propose DualNeRF, a dual-field representation with a static field for guidance signal providing and a dynamic field for flexible editing. This novel architecture provides new guidance signals to the model during IDU and results in edits with clearer background.
    \item We introduce a simulated annealing strategy into IDU, which endows our model with the ability to address local optima.
    \item We design a CLIP-based consistency indicator to measure the edits of IP2P, which can strengthen the impact of high-quality edits while weakening the low-quality ones.
    \item Experiments demonstrate that our method achieves better editing performance compared to IN2N.
\end{itemize}

\section{Related work}
\label{sec: related work}

\subsection{Text-guided Image Editing by Diffusion}
In recent years, diffusion models have become the most popular and powerful 2D image synthesis model due to their impressive generation results \cite{sohl2015deep, ho2020denoising, song2020denoising, song2022diffusion}. Combined with language models, text-guided diffusion models were proposed and achieved promising results according to user-provided captions \cite{nichol2021glide, rombach2022high, saharia2022photorealistic, ramesh2022hierarchical}. Based on these brilliant text-to-image diffusion models, diffusion-based image editing shows significant progress. Some of them finetune a pre-trained latent diffusion model (LDM) before editing \cite{zhang2022sine, valevski2022unitune, kawar2023imagic}. However, these methods consume huge computing power and suffer from low diversity. Sdedit \cite{meng2021sdedit} proposes to edit target images by first adding noise and denoising according to the prompts. In this way, no finetuning is required, but prone to over-edit. Prompt-to-prompt \cite{hertz2022prompt} demonstrates that a more relative editing result to the original image can be obtained by controlling the cross-attention map of U-Net \cite{ronneberger2015u}. Pix2pix-zero \cite{parmar2023zero} achieves similar performance with cross-attention guidance via L2 loss. Text2LIVE \cite{bar2022text2live} generates an edit RGBA layer, namely a color map and an opacity map, which blends into the original image for editing. Diffedit \cite{couairon2022diffedit} uses the difference introduced by conditions to guide localized edits, but can only deal with some relatively easy prompts. Instructpix2pix (IP2P) \cite{brooks2023instructpix2pix} synthesises a huge image-caption-image editing dataset based on GPT \cite{brown2020language}, Stable Diffusion \cite{rombach2022high} and Prompt-to-prompt \cite{hertz2022prompt}. Trained on this dataset, IP2P achieves the SOTA image editing result but still suffers from over-edit and instability. In this work, we use IP2P to edit images of different views and iteratively update the training dataset following \cite{haque2023instruct}. A CLIP-based consistency indicator is proposed to filter out low-quality edits, preventing them from contaminating the dataset.

% Null-text Inversion \cite{mokady2023null}

\subsection{Neural Radiance Field Editing}
% The editing of NeRF has become a significant problem in the field. Many early works pay attention to some specific tasks of NeRF editing. For geometry deformation, previous works \cite{yuan2022nerf, garbin2022voltemorph} build a canonical field along with a deformation field to deform the geometry of the target object. Cages or tetrahedrons can also be used to regularize the deforming process \cite{xu2022deforming, tertikas2023partnerf, peng2022cagenerf}. For texture (or color) editing \cite{liu2021editing, xiang2021neutex, gong2023recolornerf, kuang2023palettenerf}, scene manipulation \cite{niemeyer2021giraffe, xu2023discoscene}, relighting \cite{li2022climatenerf, zhang2021nerfactor, boss2021neural, srinivasan2021nerv} and stylization \cite{zhang2022arf, wang2023nerf}.

The editing of NeRF \cite{mildenhall2021nerf} has become a significant problem in the field. Many early works pay attention to some specific tasks of NeRF editing, including geometry deformation \cite{yuan2022nerf, garbin2022voltemorph, xu2022deforming, tertikas2023partnerf, peng2022cagenerf, jambon2023nerfshop}, texture or color editing \cite{liu2021editing, xiang2021neutex, gong2023recolornerf, kuang2023palettenerf, lee2023ice}, scene manipulation \cite{niemeyer2021giraffe, xu2023discoscene}, relighting \cite{li2022climatenerf, zhang2021nerfactor, boss2021neural, srinivasan2021nerv} and stylization \cite{zhang2022arf, wang2023nerf}. Despite the promising results, most of them can only deal with one or two tasks limited in their paper. Moreover, their editing operations are usually unuser-friendly.

In recent years, many researchers have utilized the powerful 2D priors in the pre-trained vision-language models \cite{radford2021learning, nichol2021glide, rombach2022high} to edit NeRF by text prompts. Clip-nerf \cite{wang2022clip} leverages a CLIP \cite{radford2021learning} image/text encoder to maintain the consistency between the text prompt and rendered results. In a similar way, NeRF-Art \cite{wang2023nerf} also chooses to use CLIP to stylize target NeRF models driven by text prompts. Instruct-NeRF2NeRF (IN2N) \cite{haque2023instruct} proposes to iteratively update the training dataset by IP2P \cite{brooks2023instructpix2pix} along with model training to counter IP2P's multi-view inconsistent editing results. Although IN2N can eventually converge into a well-looking result, it still suffers from problems of blurry backgrounds and local optima. Some followers of IN2N \cite{fang2023text, mirzaei2023watch} still unable to solve both problems. In this work, our method follows the iterative dataset update (IDU) strategy of IN2N, while proposing a novel dual-field network architecture to address the aforementioned problems.

\section{Preliminaries}

\begin{figure*}[t]
    \centering
    \includegraphics[width=1.0\linewidth]{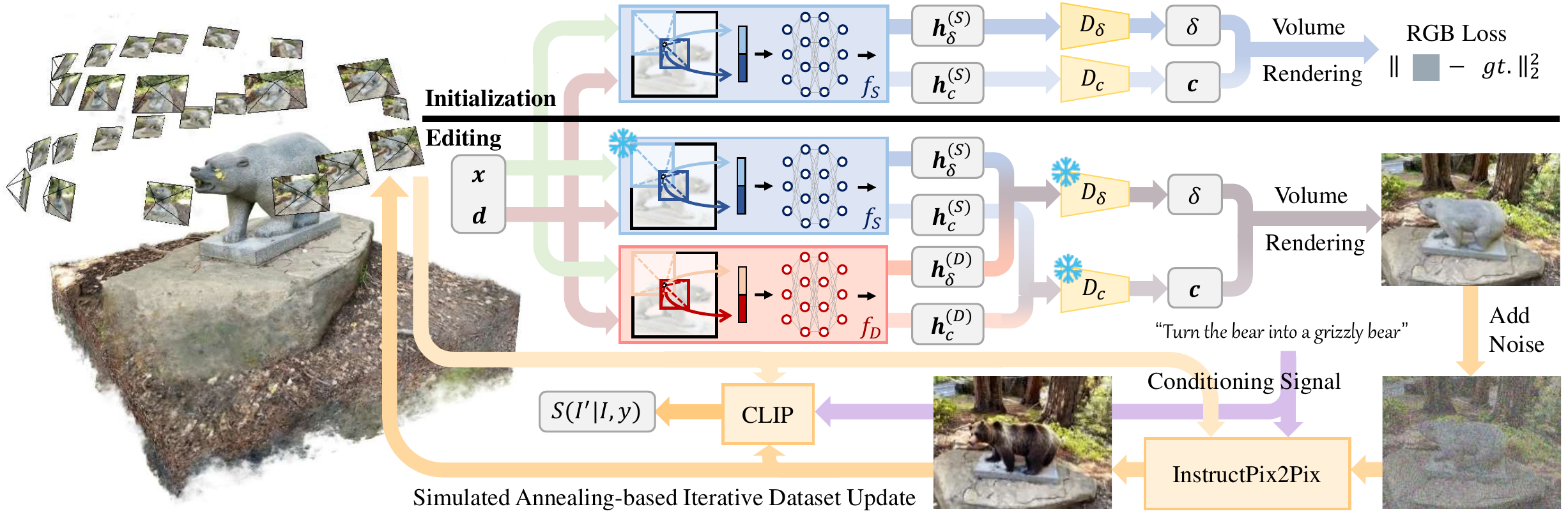}
    \caption{\textbf{The Overview of DualNeRF.} DualNeRF consists of two neural radiance fields, including a static field $f_S$ and a dynamic field $f_D$ with the same network architecture. The static field $f_S$ is trained in the field initialization stage and frozen in the editing stage. The dynamic field $f_D$ is enabled during the editing stage and trained to achieve field editing. Two fields fuse in the hidden feature level. A simulated annealing-based IDU strategy is used to perform editing. Furthermore, a CLIP-based consistency indicator is calculated based on the inputs and outputs, which filters out low-quality edits softly and therefore cleans up the updated dataset.}
    \label{fig: overview}
\end{figure*}

\subsection{Neural radiance fields}
\label{sec: Neural radiance fields}
NeRFs \cite{mildenhall2021nerf} represent a target scene/object implicitly by neural networks (NNs). Specifically, given a space position $\mathbf{x} = (x, y, z)$ and a view direction $\mathbf{d} = (\theta, \phi)$, a NeRF model $f$ outputs the occupancy $\sigma(\mathbf{x})$ at $\mathbf{x}$ and the radiance $\mathbf{c}(\mathbf{x}, \mathbf{d})$ at $\mathbf{x}$ viewed from $\mathbf{d}$, namely
\begin{equation}
    (\sigma(\mathbf{x}), \mathbf{c}(\mathbf{x}, \mathbf{d})) = f(\mathbf{x}, \mathbf{d})
    \label{equ: NeRF}
\end{equation}
The rendering of NeRF can be achieved by volume rendering. For a ray $\mathbf{r} = \mathbf{o} + t\mathbf{d}$, where $\mathbf{o}$ is the origin and $\mathbf{d}$ is the direction, $N$ samples $\{x_i = \mathbf{o} + t_i\mathbf{d}\}_{i=1}^{N}$ are sampled along the ray. The color $\hat{C}(\mathbf{r})$ of the ray is calculated as an alpha blending: $\hat{C}(\mathbf{r}) = \sum_{i=1}^{N}w_i\mathbf{c}(\mathbf{x}_i, \mathbf{d})$,
% \begin{equation}
%     \hat{C}(\mathbf{r}) = \sum_{i=1}^{N}w_ic(\mathbf{x}_i, \mathbf{d})
%     \label{equ: vol. rendering 1}
% \end{equation}
where $w_i = T_i(1-\exp(-\delta_i\sigma(\mathbf{x}_i))$
% \begin{equation}
%     w_i = \exp(-\sum_{j=1}^{i-1}\delta_j\sigma(\mathbf{x}_j))(1-\exp(-\delta_i\sigma(\mathbf{x}_i))
%     \label{equ: vol. rendering 2}
% \end{equation}
is the blending weight of $\mathbf{c}(\mathbf{x}_i, \mathbf{d})$. $\delta_i = t_{i+1} - t_i$ is the distance between adjacent sample points. $T_i = \exp(-\sum_{j=1}^{i-1}\delta_j\sigma(\mathbf{x}_j))$ is the transmittance.

The differentiable nature of the volume rendering helps NeRF to be trained by stochastic gradient descent. Specifically, given a multi-view dataset $\mathcal{I} = \{(I_j, P_j)\}_{j=1}^{N}$, where $I_j$ is a ground truth image and $P_j$ represents its camera pose, an L2-loss between the rendering $\hat{C}(\mathbf{r})$ and ground truth $C(\mathbf{r})$ can be used to train the NeRF model:
\begin{equation}
    L_{rgb} = \sum_{\mathbf{r} \in \mathcal{R}}||\hat{C}(\mathbf{r}) - C(\mathbf{r})||^2_2
    \label{equ: nerf loss}
\end{equation}
where $\mathcal{R}$ is a batch of rays sampled from $\mathcal{I}$. Additional losses such as LPIPS loss \cite{8578166} can also be used to improve the rendering quality.

\subsection{Instruct-NeRF2NeRF}
\label{sec: Instruct-NeRF2NeRF}
Instruct-NeRF2NeRF \cite{haque2023instruct} (IN2N) proposes a new NeRF editing framework called ``Iterative dataset update" (IDU). Specifically, the framework contains two steps:
\begin{enumerate}
    \item A dataset updating step,
    % in which $K$ renderings $\hat{\mathbf{I}} = \{\hat{I}_i\}_{i=1}^{K}$ rendered from $K$ random views $\mathbf{P} = \{P_i\}_{i=1}^{K}$ in the training dataset $\mathcal{I}$ are edited by an IP2P model conditioned on a prompt $\mathbf{y}$ and the original image $I_i$, followed by replacing the corresponding images in $\mathcal{I}$, as equation \ref{equ: update dataset} shows:
    % \begin{equation}
    %     \mathcal{I}' = \{(\text{IP2P}(\hat{I}_i | I_i, \mathbf{y}), P_i) | P_i \in \mathbf{P}\} \cup \{(I_j, P_j) \in \mathcal{I} | P_j \notin \mathbf{P} \}
    %     \label{equ: update dataset}
    % \end{equation}
    % where $\mathcal{I}'$ is the updated dataset.
    in which $d$ images in the training dataset $\mathcal{I}$ are replaced by $d$ edits $\{I'_i\}_{i=1}^{d}$ generated by a text-driven image editing model conditioned on the prompt $\mathbf{y}$ and the original image $\{I_i\}_{i=1}^{d}$, resulting in an updated dataset $\mathcal{I}'$ mixed with old and new images.
    \item A NeRF updating step, where the NeRF model is trained on the new dataset $\mathcal{I}'$ for $n$ iterations.
\end{enumerate}
These two steps alternate until convergence. The image editing model used in the first step by IN2N is InstructPix2pix \cite{brooks2023instructpix2pix} (IP2P), a SOTA text-driven image editing method based on Stable Diffusion \cite{rombach2022high}.

\section{Method}
In this work, we introduce DualNeRF, a system aiming at editing a target scene complying with a user-provided prompt $y$.
Following IN2N \cite{haque2023instruct}, we start with reconstructing the target scene by a NeRF given a dataset of multiview images along with corresponding cameras $\mathcal{I} = \{I_j, P_j\}_{j=1}^{N}$, and then edit the scene based on IDU \cite{haque2023instruct} strategy.

In this section, we first present an overview introduction of the dual-field representation of DualNeRF in Sec. \ref{sec: Dual-field Representation}.
After that, we introduce how to combine simulated annealing (SA) strategy \cite{kirkpatrick1983optimization} into the pipeline of IDU to mitigate the problem of local optima in Sec. \ref{sec: Simulated Annealing Strategy}.
We further use a consistency indicator based on CLIP \cite{radford2021learning} to filter out low-quality edits of IP2P therefore further improving the editing results in Sec. \ref{sec: Editing Result Filtering}.
Implementation details of our model are shown in Sec. \ref{sec: Implementation Details}.

\subsection{Dual-field Representation}
\label{sec: Dual-field Representation}
% We design a dual-field representation to stabilize the features of unedited regions during training. There are two neural networks representing two fields contained in DualNeRF, including a major field $f_S$ and an editing field $f_D$.
DualNeRF contains two neural networks. One is set as the static field $f_S$, whose parameters are trained to faithfully reconstruct the original scene and frozen during editing for guidance signal providing. Another is designed as the dynamic field $f_D$, which is gradually enabled and trained during editing. Hidden features from the two fields fused by decoders output the final results. An overview of DualNeRF is shown in Fig. \ref{fig: overview}.

\paragraph{Field Initialization.}
Given dataset $\mathcal{I}$, we first train the static field $f_S$ to reconstruct the original scene for the initialization of the following editing stage. Modified from Equ. \ref{equ: NeRF}, the output of $f_S$ are two hidden features:
\begin{equation}
    (\mathbf{h}_{\delta}^{(S)}, \mathbf{h}_{c}^{(S)}) = f_S(\mathbf{x}, \mathbf{d})
\end{equation}
where $\mathbf{h}^{(S)}_{\sigma}$ denotes a density feature and $\mathbf{h}^{(S)}_{c}$ denotes a color feature. These two features are further decoded into $(\sigma, \mathbf{c})$ by a density decoder $D_{\sigma}$ and a color decoder $D_{c}$ respectively.
% Thanks to the huge progress in NeRF, only one field $f_S$ is enough to reconstruct the target scene faithfully.
A well-trained $f_S$ has two benefits: (1) It gives a good initialization of the following editing stage; (2) It stores authentic information about the original scene, which can be used as guidance to stabilize the editing process.

\paragraph{Field Editing.}
In the editing stage, a new dynamic field $f_D$ with the same architecture as $f_S$ is introduced into the model.
Given a query point $\mathbf{x}$ and viewing direction $\mathbf{d}$, these two variables are sent into both fields, resulting in two pairs of hidden features: $(\mathbf{h}_{\delta}^{(*)}, \mathbf{h}_{c}^{(*)})$, where $* \in \{S, D\}$.
The fusion of $f_S$ and $f_D$ is achieved by weighted sums of these hidden features:
\begin{align}
\mathbf{h}_{\delta} = (1 - w_{\delta})\mathbf{h}_{\delta}^{(S)} + w_{\delta}\mathbf{h}_{\delta}^{(D)} \\
\mathbf{h}_{c} = (1 - w_{c})\mathbf{h}_{c}^{(S)} + w_{c}\mathbf{h}_{c}^{(D)}
\end{align}
where $w_{\delta}, w_{c} \in [0, 1]$ are the blending weights controlling the editing intensity brought from $f_D$. 

During the editing stage, the parameters of $f_S$ are frozen to preserve features of the original scene, while the parameters of $f_D$ are trained to implement edits matched with prompt $y$. We gradually increase the values of $w_{\delta}$ and $w_{c}$ during editing, allowing the model to transfer from the original scene to the edited version smoothly.

More concretely, $w_{\delta}$ and $w_{c}$ are set to $0$ at the beginning of the editing stage to initialize the editing process as the original scene. We increase them in a tanh formula up to upper bounds $w_{\delta}^{max}$ and $w_{c}^{max}$, as Equ. \ref{equ: w} shows:
\begin{equation}
    w_{*} = w_{*}^{max}\tanh(\lambda t)
    \label{equ: w}
\end{equation}
where $* \in \{\delta, c\}$. $\lambda$ is a hyperparameter controlling the growing velocity of $w_*$ and $t$ represents the iteration number. Note that when the upper bounds are set to $0$, the scene remains as the original version without editing. When the upper bounds are set to $1$, the intensity of $f_S$ will gradually fade away after sufficiently long iterations.

In practice, the value of the $w_{\delta}^{max}$ and $w_{c}^{max}$ are always set to less than $1$. In this way, the abundant and authentic features of the original scene stored in $f_S$ can be held and leveraged during iterations. These features serve as ``anchors" to guide the training process not drifting far away from the initial state, mitigating the impact of color jitters in unedited areas brought by IP2P's edits with distorted backgrounds. As a result, the editing progresses more stably, achieving local editing with a clearer and more restored background.

\subsection{Simulated Annealing Strategy}
\label{sec: Simulated Annealing Strategy}
As we mentioned in Sec. \ref{sec: Instruct-NeRF2NeRF}, IN2N is prone to fall into local optima due to the mutual promotion between IP2P edits and NeRF optimization when facing artifacts. To solve this problem, a simulated annealing strategy is plugged into the pipeline of IDU.

Specifically, we randomly jitter the value of $w_{\delta}$ and $w_c$ by multiplying a random scaler $\gamma \in [0, 1]$ in each dataset updating step. Note that if $\gamma = 1$, the model remains the latest version, while if $\gamma = 0$, the model retreats to the original scene. We randomly accept to render from a retreated model with $\gamma < 1$ with probability
\begin{equation}
    p(\gamma) = \exp(\frac{\gamma - 1}{T_t})
\end{equation}
where $T_t$ is the temperature in iteration $t$ with a logarithmic decaying expression starting from an initial temperature $T_0$:
\begin{equation}
    T_t = \frac{T_0}{\lg(10 + t)}
\end{equation}

Rendering from a retreated model outputs a more natural result when artifacts should have appeared. This will be more friendly to IP2P as IP2P is trained on a high-quality dataset with few artifacts. In this way, our simulated annealing strategy endows the model with the ability to address the issue of local optima.

\begin{figure}[t]
    \centering
    \includegraphics[width=1.0\linewidth]{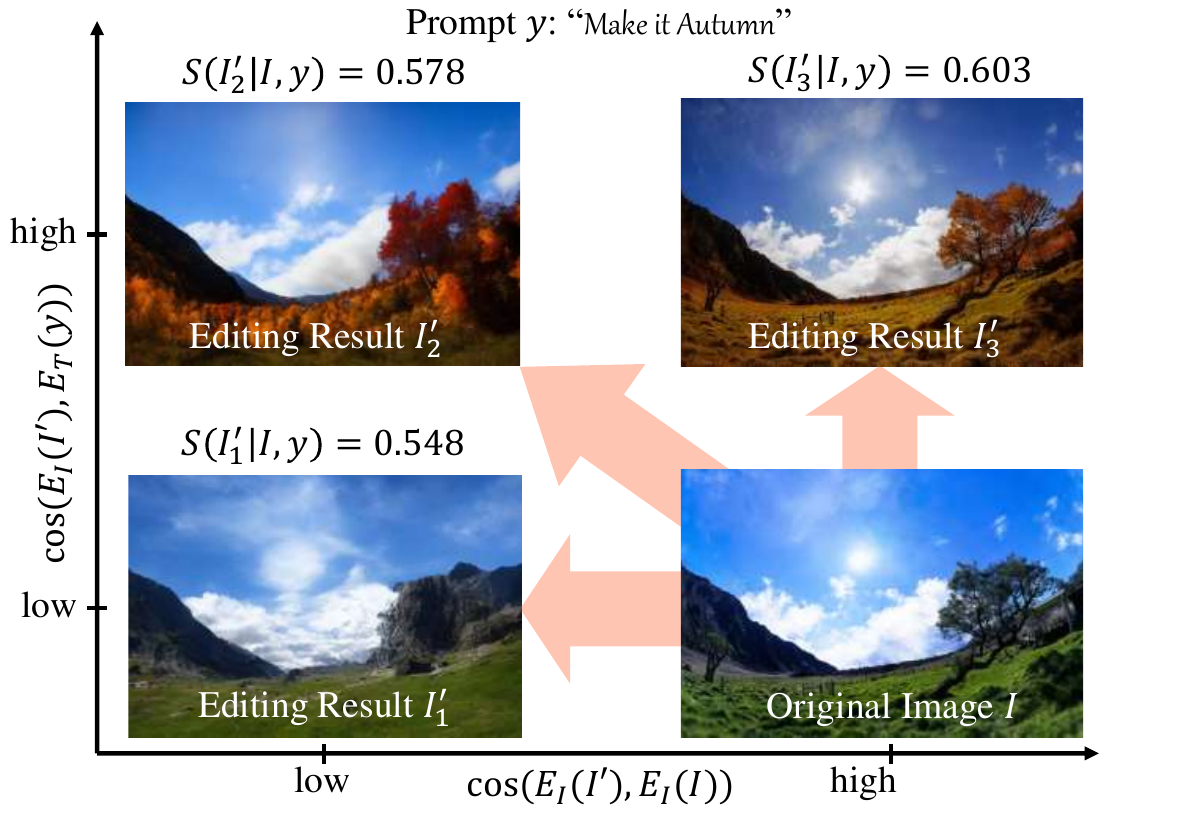}
    \caption{\textbf{Edits with Their CLIP-based Consistency.} The right bottom image is the original image $I$, while the rest images are three IP2P edits based on the prompt "Make it Autumn". $I'_1$ is inconsistent with both original image $I$ and the prompt $y$, which leads to the lowest consistency score $\mathcal{S}$. $I'_2$ transfers the original image to an Autumn scenery but fails to restore the original image. $I'_3$ is the best edit with high consistency to both $I$ and $y$, resulting in the highest $\mathcal{S}$. These examples demonstrate the ability of $\mathcal{S}$ to filter out low-quality edits.}
    \label{fig: clip consistency}
\end{figure}

\subsection{Editing Result Filtering}
\label{sec: Editing Result Filtering}
A CLIP-based consistency indicator $\mathcal{S}$ is further used to measure the editing quality of an editing result $I'$. Given the original image $I$ and a prompt $y$, the CLIP-based consistency of $I'$ is defined as Equ. \ref{equ: clip}:
\begin{equation}
    \mathcal{S}(I' | I, y) = \cos(E_{I}(I'), E_{I}(I)) \cdot \cos(E_{I}(I'), E_{T}(y))
    \label{equ: clip}
\end{equation}
% \begin{equation}
%     \mathcal{S}(I' | y) = \frac{<E_{I}(I'), E_{I}(I)>}{||E_{I}(I')||_2 \cdot ||E_{I}(I)||_2} \cdot \frac{<E_{I}(I'), E_{T}(y)>}{||E_{I}(I')||_2 \cdot ||E_{T}(y)||_2}
%     \label{equ: clip}
% \end{equation}
% \begin{equation}
%     \mathcal{S}(I' | y) = <E_{I}(I'), E_{I}(I)> \cdot <E_{I}(I'), E_{T}(y)>
%     \label{equ: clip}
% \end{equation}
where $\cos(\cdot, \cdot)$ denotes the cosine similarity (normalized within $[0, 1]$) between two vectors.
$E_{I}$ and $E_{T}$ are the image encoder and text encoder of CLIP.
This consistency indicator considers the consistency between both the edited image $I'$ with the original image $I$ and $I'$ with the text prompt $y$.
Only edits satisfying both consistencies simultaneously obtain a high value of $\mathbf{S}$.
Examples of edits with different consistency indicators $\mathbf{S}$ are shown in Fig. \ref{fig: clip consistency}.
% where $<\cdot, \cdot>$ denotes the inter product operation between two vectors.
% $E_{I}$ and $E_{T}$ are the image encoder and text encoder of CLIP. This consistency indicator measures the consistency between the edited image $I'$ with the text prompt $y$, where higher values of $S$ indicate higher obedience towards $y$ by $I'$.
We use $\mathcal{S}$ to regulate the intensity of loss calculated from rays in the corresponding image. Specifically, the loss function in Equ. \ref{equ: nerf loss} is modified as follows:
\begin{equation}
    L_{rgb} = \sum_{I_i \in \mathcal{I}}\sum_{\mathbf{r} \in \mathcal{R}_i}\frac{\mathcal{S}(I_i' | I_i, y)}{\bar{\mathcal{S}}}||\hat{C}(\mathbf{r}) - C(\mathbf{r})||^2_2
    \label{equ: nerf loss new}
\end{equation}
where $\mathcal{R}_i$ represents the rays sampled from image $I_i$ in the current batch. $\bar{\mathcal{S}}$ is the mean value of all $\mathbf{S}$'s of different views for normalization. In this way, editing results with higher consistency with the original image and input prompt contribute more to the training of the model than the low-quality ones. This process can be seen as a dataset-cleaning operation. Note that the value of $S$ will be cached and used until updated in the next round.

% Note that we do not consider the consistency between $I'$ and the original image $I$ in $\mathbf{S}$, since we find that the cosine similarity between their CLIP features highly depends on the viewing angle. Empirically, frontal views usually have higher CLIP-based consistency between $I'$ and $I$ than the rear views, even if the edits in rear views have higher visual quality. This leads to a bias towards edits in frontal views, which goes against our expectations.

\subsection{Implementation Details}
\label{sec: Implementation Details}
The architecture of the two fields is implemented by \textit{nerfacto} provided in NeRFStudio \cite{nerfstudio} due to its high effectiveness and efficiency. The upper bounds of the blending weights are set to $w^{max}_{\delta} = w^{max}_c = 0.1$, which controls the maximum strength of the modification brought by the dynamic field $f_D$ at a moderate intensity. The growing velocity $\lambda$ of $w_*$ takes the value of $0.005$, which leads to an appropriate speed to smoothly introduce $f_D$ into the model. The two decoders $D_{\sigma}$ and $D_{c}$ are empirically designed as two activation functions, namely a truncated exponential function and a sigmoid function, which cause minimal impacts to the hidden features while being strong enough to merge features in a meaningful way. More details can be seen in the supplementary material. During the IDU process with our simulated annealing strategy, the temperature of SA is initialized as $T_0 = 1$ and dropped in a logarithmic way. The training of our model uses both the RGB loss in Equ. \ref{equ: nerf loss} and LPIPS loss \cite{8578166}. The model is optimized for $15k$ iterations for editing, which takes about $1$ hour on a single NVIDIA GeForce RTX 3090. Other hyperparameters follow the settings in \cite{haque2023instruct} for a fair comparison. The code will be released later.

\section{Experiments}

\begin{figure*}
    \centering

    \begin{subfigure}{0.33\linewidth}
        \begin{minipage}[t]{1.0\linewidth}
            \centering
            \includegraphics[width=1.0\linewidth]{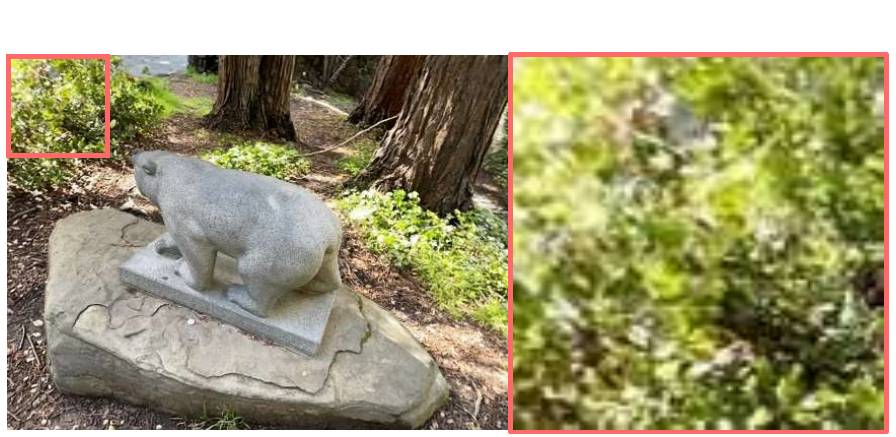}
            % \caption{Original Scene}
        \end{minipage}
    \end{subfigure}
    \hfill
    \begin{subfigure}{0.33\linewidth}
        \begin{minipage}[t]{1.0\linewidth}
            \centering
            \includegraphics[width=1.0\linewidth]{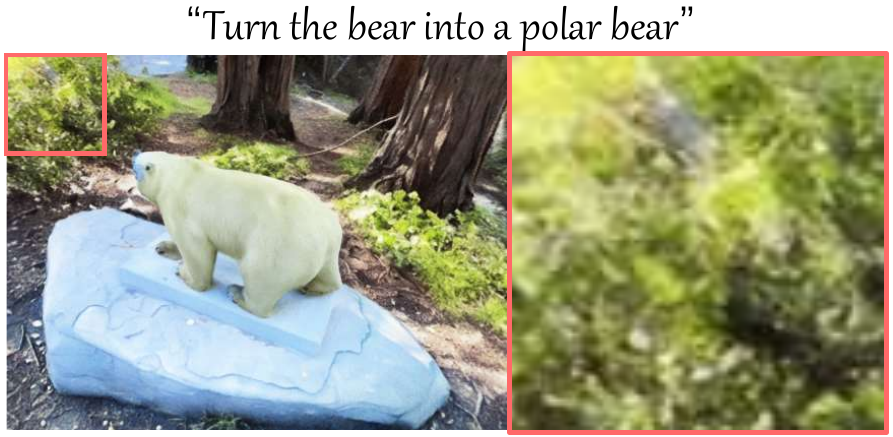}
            % \caption{Instruct-NeRF2NeRF}
        \end{minipage}
    \end{subfigure}
    \hfill
    \begin{subfigure}{0.33\linewidth}
        \begin{minipage}[t]{1.0\linewidth}
            \centering
            \includegraphics[width=1.0\linewidth]{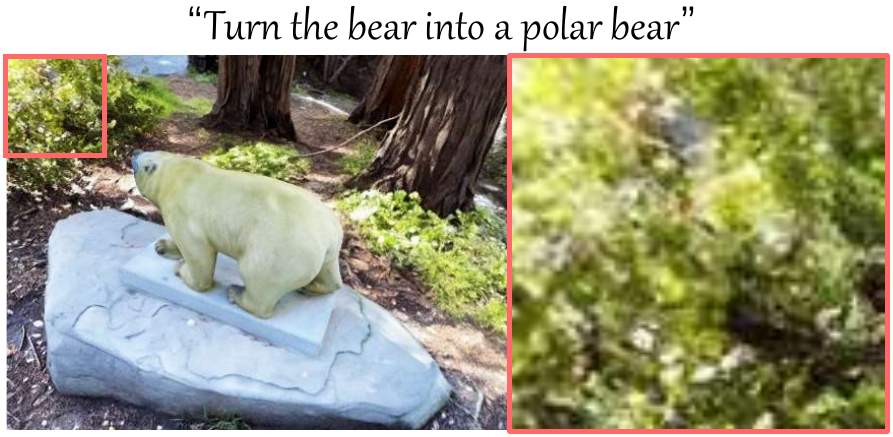}
            % \caption{DualNeRF}
        \end{minipage}
    \end{subfigure}

    \begin{subfigure}{0.33\linewidth}
        \begin{minipage}[t]{1.0\linewidth}
            \centering
            \includegraphics[width=1.0\linewidth]{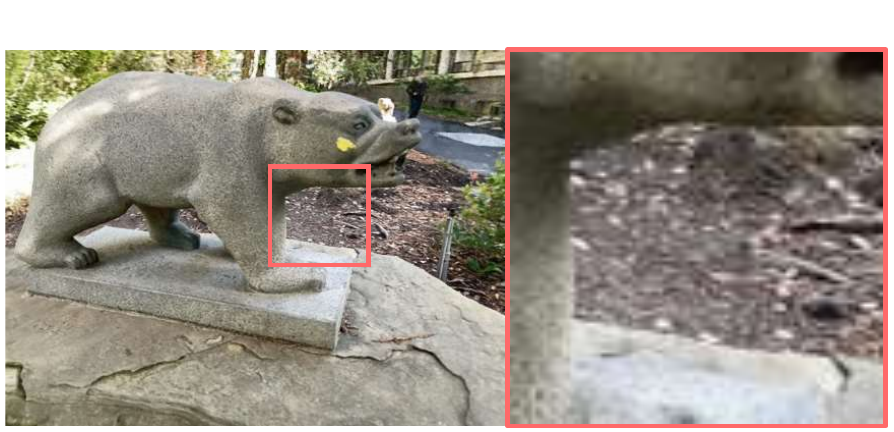}
            % \caption{Original Scene}
        \end{minipage}
    \end{subfigure}
    \hfill
    \begin{subfigure}{0.33\linewidth}
        \begin{minipage}[t]{1.0\linewidth}
            \centering
            \includegraphics[width=1.0\linewidth]{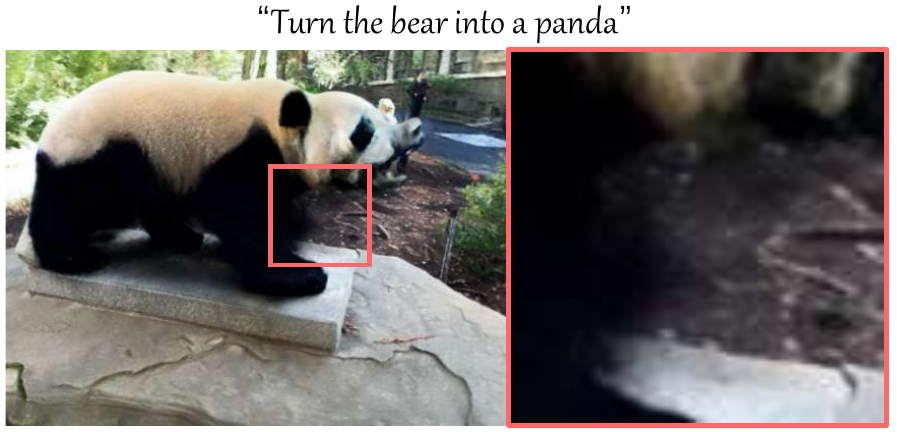}
            % \caption{Instruct-NeRF2NeRF}
        \end{minipage}
    \end{subfigure}
    \hfill
    \begin{subfigure}{0.33\linewidth}
        \begin{minipage}[t]{1.0\linewidth}
            \centering
            \includegraphics[width=1.0\linewidth]
            {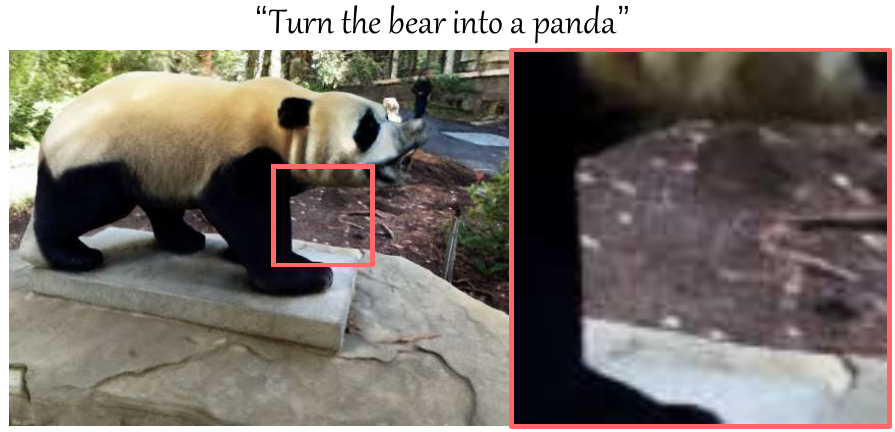}
            % \caption{DualNeRF}
        \end{minipage}
    \end{subfigure}

    \begin{subfigure}{0.33\linewidth}
        \begin{minipage}[t]{1.0\linewidth}
            \centering
            \includegraphics[width=1.0\linewidth]{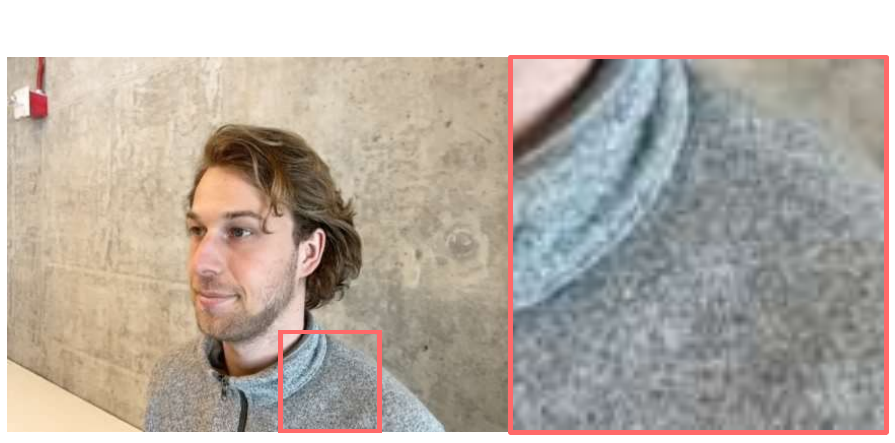}
            % \caption{Original Scene}
        \end{minipage}
    \end{subfigure}
    \hfill
    \begin{subfigure}{0.33\linewidth}
        \begin{minipage}[t]{1.0\linewidth}
            \centering
            \includegraphics[width=1.0\linewidth]{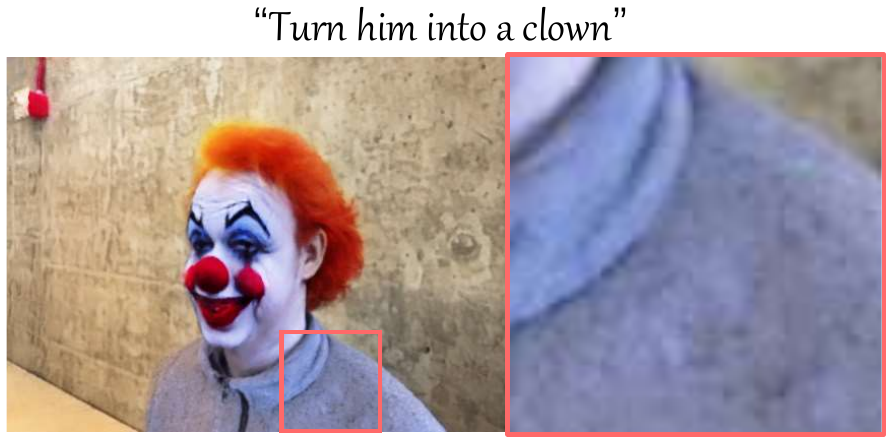}
            % \caption{Instruct-NeRF2NeRF}
        \end{minipage}
    \end{subfigure}
    \hfill
    \begin{subfigure}{0.33\linewidth}
        \begin{minipage}[t]{1.0\linewidth}
            \centering
            \includegraphics[width=1.0\linewidth]{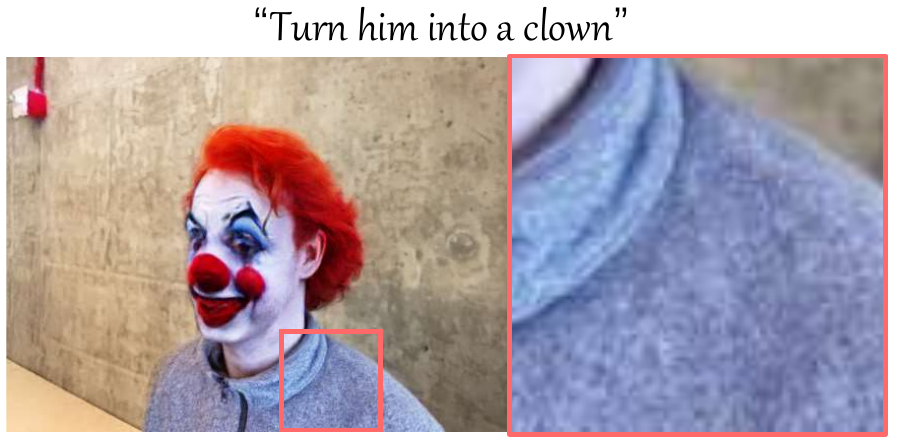}
            % \caption{DualNeRF}
        \end{minipage}
    \end{subfigure}

    \begin{subfigure}{0.33\linewidth}
        \begin{minipage}[t]{1.0\linewidth}
            \centering
            \includegraphics[width=1.0\linewidth]{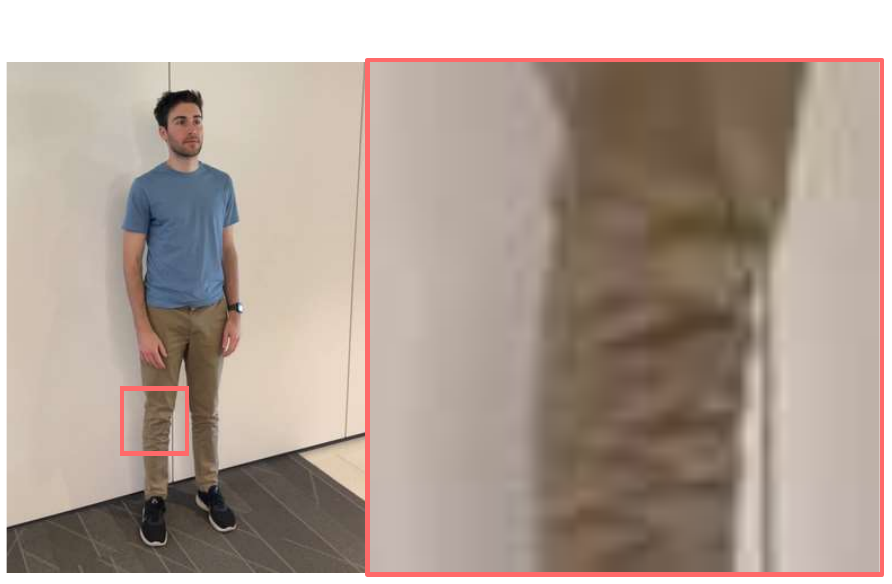}
            % \caption{Original Scene}
        \end{minipage}
    \end{subfigure}
    \hfill
    \begin{subfigure}{0.33\linewidth}
        \begin{minipage}[t]{1.0\linewidth}
            \centering
            \includegraphics[width=1.0\linewidth]{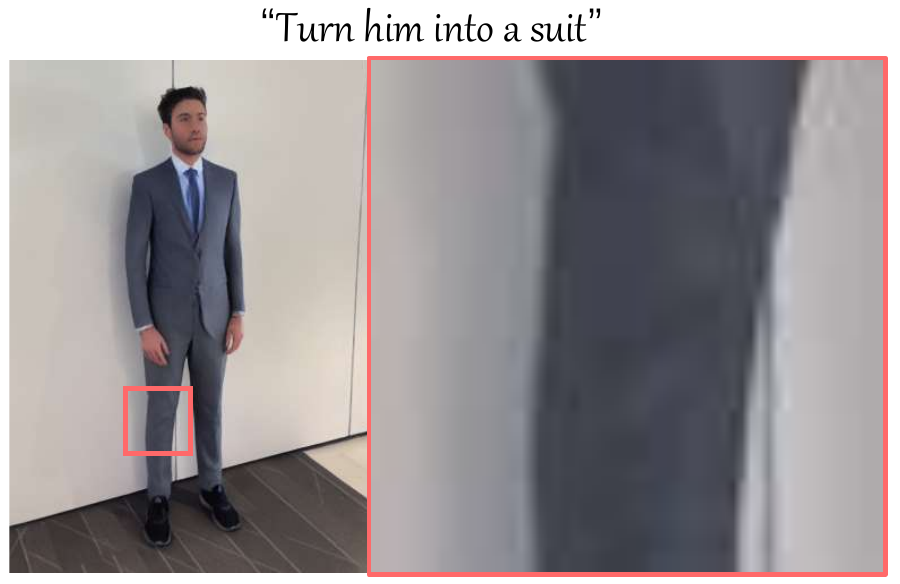}
            % \caption{Instruct-NeRF2NeRF}
        \end{minipage}
    \end{subfigure}
    \hfill
    \begin{subfigure}{0.33\linewidth}
        \begin{minipage}[t]{1.0\linewidth}
            \centering
            \includegraphics[width=1.0\linewidth]{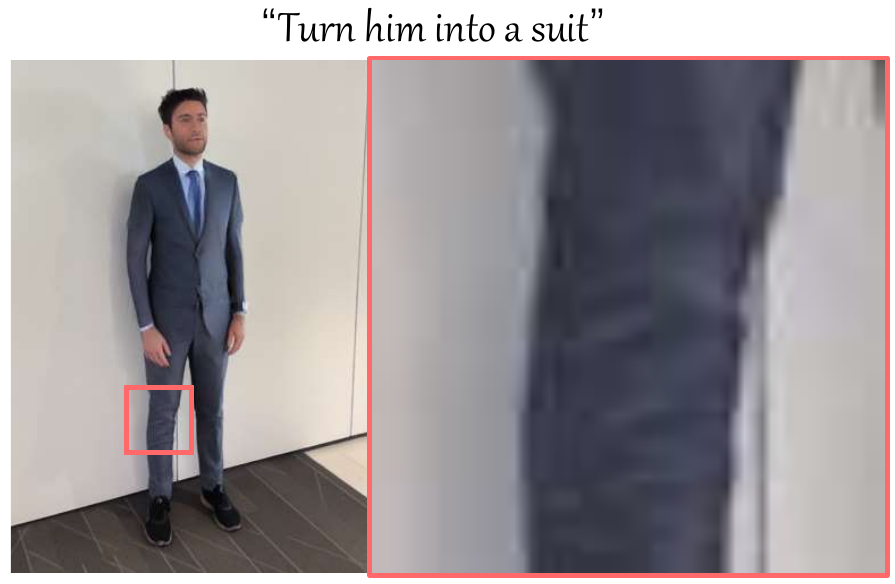}
            % \caption{DualNeRF}
        \end{minipage}
    \end{subfigure}

    \begin{subfigure}{0.33\linewidth}
        \begin{minipage}[t]{1.0\linewidth}
            \centering
            \includegraphics[width=1.0\linewidth]{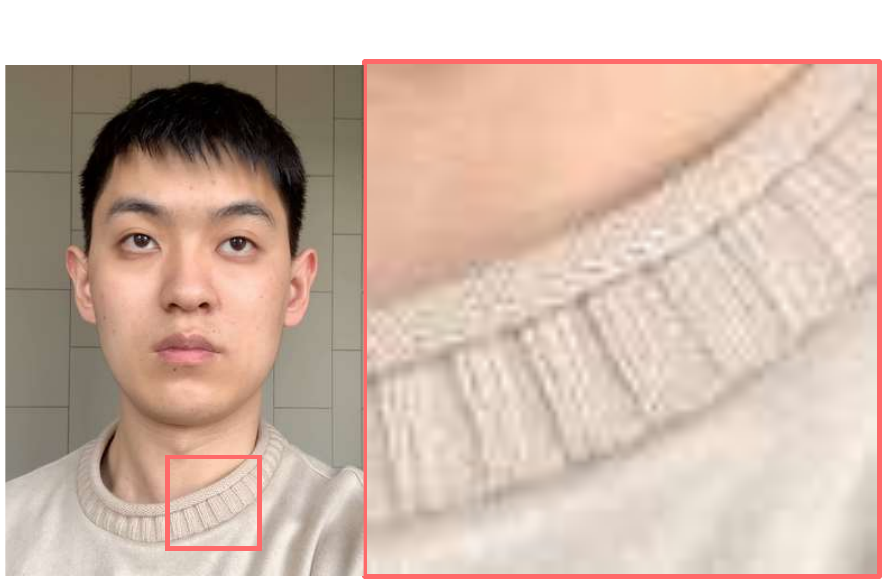}
            % \caption{Original Scene}
        \end{minipage}
    \end{subfigure}
    \hfill
    \begin{subfigure}{0.33\linewidth}
        \begin{minipage}[t]{1.0\linewidth}
            \centering
            \includegraphics[width=1.0\linewidth]{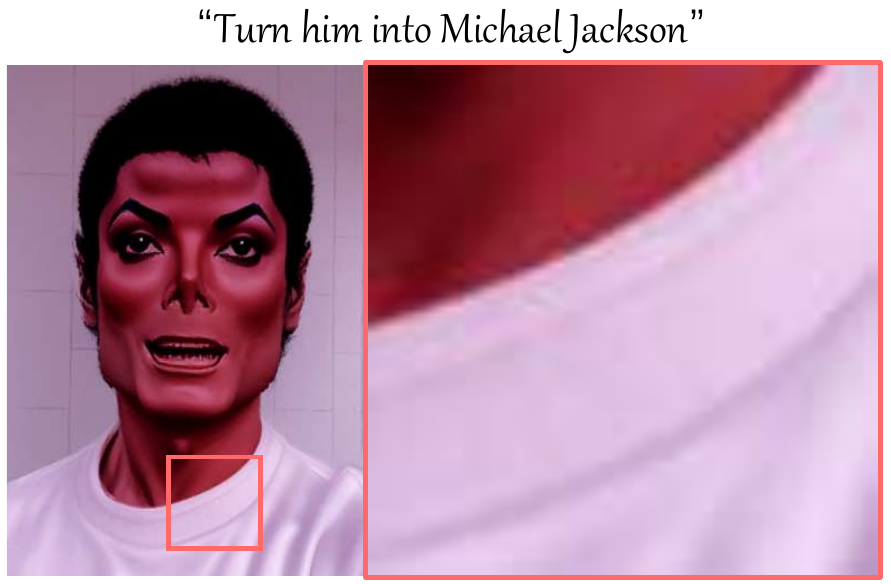}
            % \caption{Instruct-NeRF2NeRF}
        \end{minipage}
    \end{subfigure}
    \hfill
    \begin{subfigure}{0.33\linewidth}
        \begin{minipage}[t]{1.0\linewidth}
            \centering
            \includegraphics[width=1.0\linewidth]{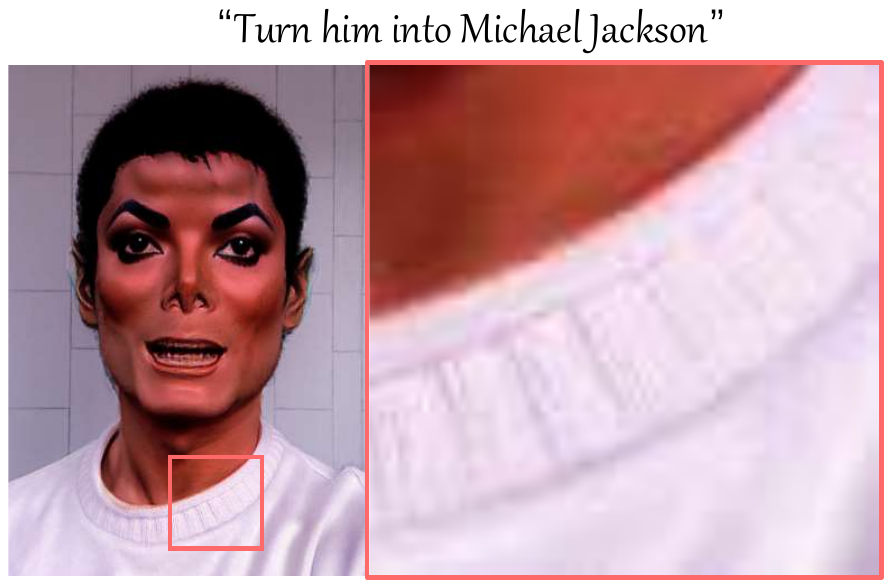}
            % \caption{DualNeRF}
        \end{minipage}
    \end{subfigure}

    \begin{subfigure}{0.33\linewidth}
        \begin{minipage}[t]{1.0\linewidth}
            \centering
            \includegraphics[width=1.0\linewidth]{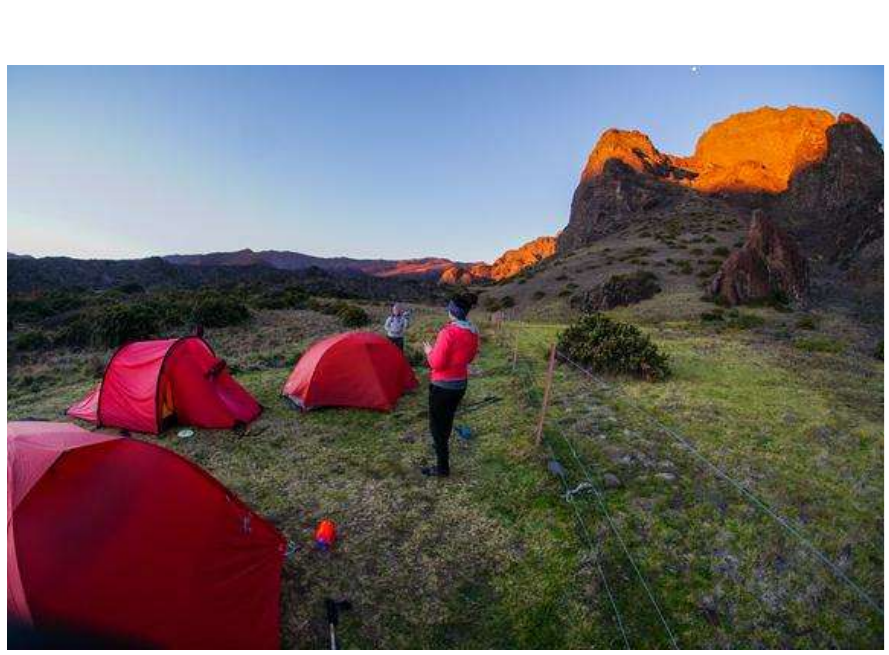}
            \caption{Original Scene}
        \end{minipage}
    \end{subfigure}
    \hfill
    \begin{subfigure}{0.33\linewidth}
        \begin{minipage}[t]{1.0\linewidth}
            \centering
            \includegraphics[width=1.0\linewidth]{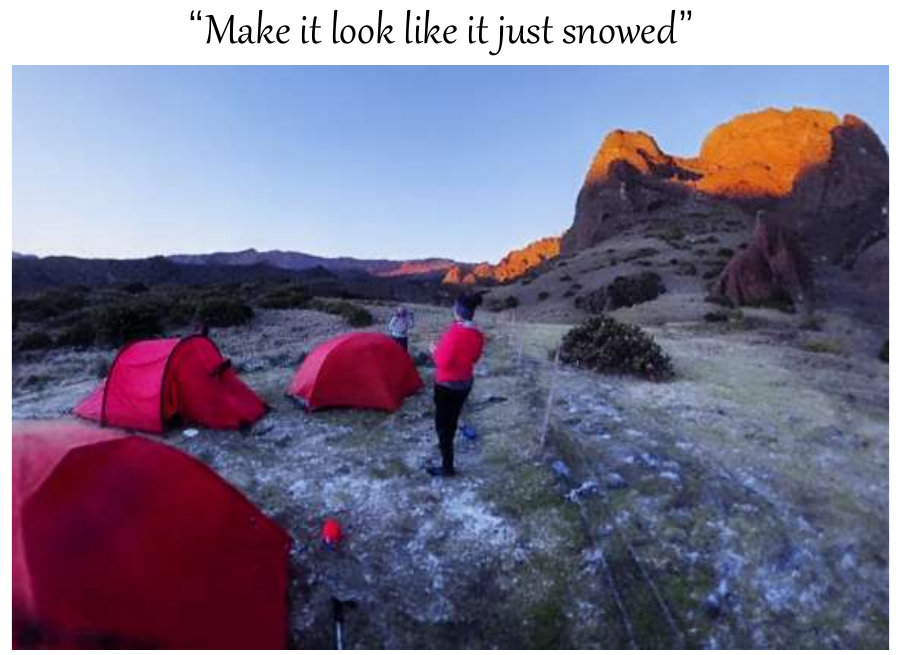}
            \caption{Instruct-NeRF2NeRF}
        \end{minipage}
    \end{subfigure}
    \hfill
    \begin{subfigure}{0.33\linewidth}
        \begin{minipage}[t]{1.0\linewidth}
            \centering
            \includegraphics[width=1.0\linewidth]{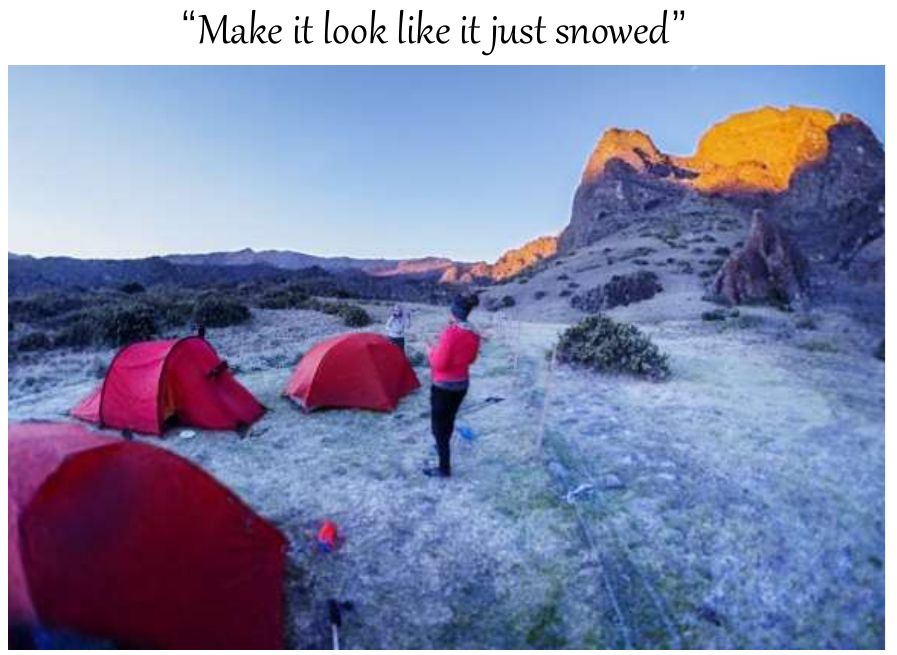}
            \caption{DualNeRF}
        \end{minipage}
    \end{subfigure}
  
    \caption{\textbf{Qualitative Results.} Comparison between DualNeRF and Instruct-NeRF2NeRF \cite{haque2023instruct} over different scenes with different prompts. Three columns respectively represent the original scene, the editing results of IN2N, and the editing results of DualNeRF. We strongly recommend readers to zoom in for a clearer observation.}
    \label{fig: qualitative results}
\end{figure*}

\subsection{Experimental Setups}
\paragraph{Datasets.}
% We conduct experiments based on scenes from IN2N \cite{haque2023instruct} and InstantNGP \cite{muller2022instant},
We conduct experiments based on scenes from IN2N \cite{haque2023instruct}, which contain $50 \sim 350$ high-quality images in various scenes, usually natural scenery or front views of a person. Following the advice of \cite{haque2023instruct}, images in the dataset are resampled to a resolution of around $512$ to match the best input resolution of IP2P \cite{brooks2023instructpix2pix}. COLMAP \cite{schoenberger2016sfm} is used to extract camera poses from images. The text prompts used in experiments are all ordinary natural languages, such as \textit{``Turn the bear into a panda"}, just like IP2P and IN2N do. 

\paragraph{Evaluation Criteria.}
\label{sec: Evaluation Criteria}
Following \cite{haque2023instruct}, we report the CLIP text-image direction similarity $C_{t2i}$ to measure the alignment between the final renderings with the prompt and CLIP direction consistency $C_{dir}$ to measure the consistency between adjacent renderings in the CLIP space.
Besides, structural similarity index (SSIM) \cite{wang2004image} is also used to measure the similarity between the original images and their edits, indicating the degree of background maintenance to some extent.
% A user study is also conducted to give a more subjective comparison by humans.

\paragraph{Baselines.}
We compare DualNeRF with SOTA 2D and 3D editing methods, including (1) Instruct-Nerf2NeRF \cite{haque2023instruct}, the SOTA 3D scene editing method based on IDU with released code; (2) InstructPix2Pix \cite{brooks2023instructpix2pix}, the underlying text-driven image editing model used in our method; (3) ControlNet \cite{zhang2023adding}, a SOTA diffusion-based image generation model controlled by signals of various modalities.

\begin{figure}
    \centering

    \begin{subfigure}{0.24\linewidth}
        \begin{minipage}[t]{1.0\linewidth}
            \centering
            \caption{Original}
            \includegraphics[width=1.0\linewidth]{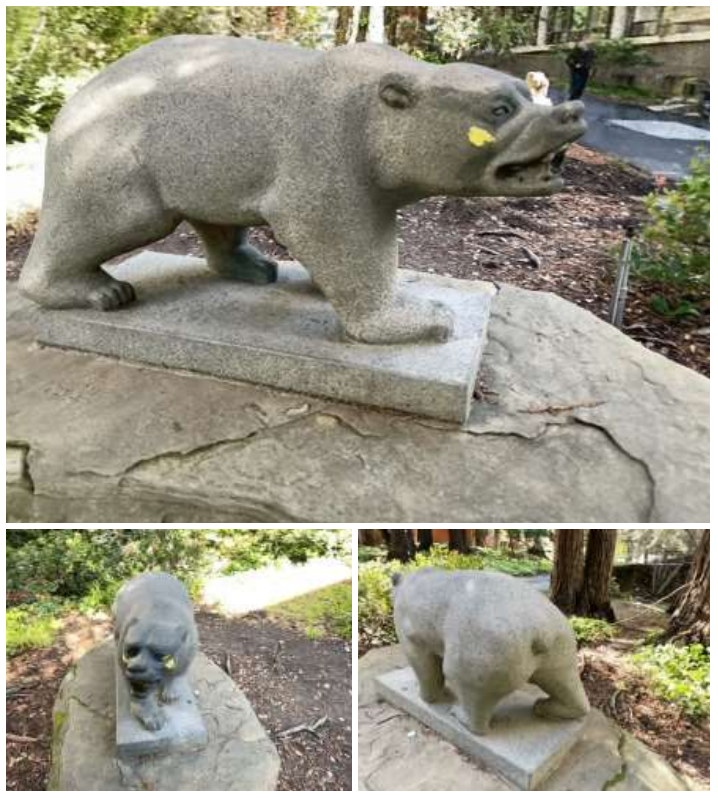}
        \end{minipage}
    \end{subfigure}
    \hfill
    \begin{subfigure}{0.24\linewidth}
        \begin{minipage}[t]{1.0\linewidth}
            \centering
            \caption{ControlNet}
            \includegraphics[width=1.0\linewidth]{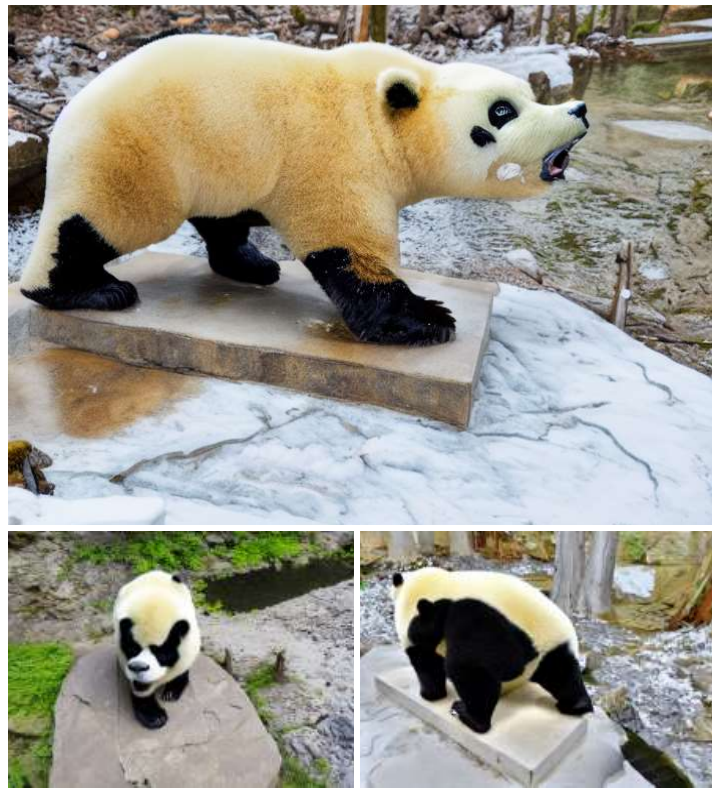}
        \end{minipage}
    \end{subfigure}
    \hfill
    \begin{subfigure}{0.24\linewidth}
        \begin{minipage}[t]{1.0\linewidth}
            \centering
            \caption{IP2P}
            \includegraphics[width=1.0\linewidth]{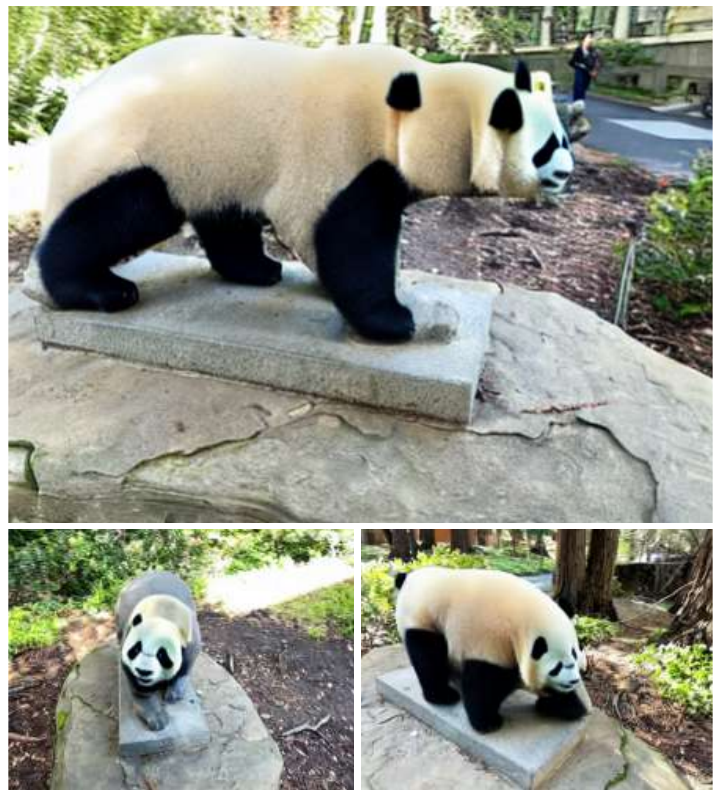}
        \end{minipage}
    \end{subfigure}
    \hfill
    \begin{subfigure}{0.24\linewidth}
        \begin{minipage}[t]{1.0\linewidth}
            \centering
            \caption{DualNeRF}
            \includegraphics[width=1.0\linewidth]{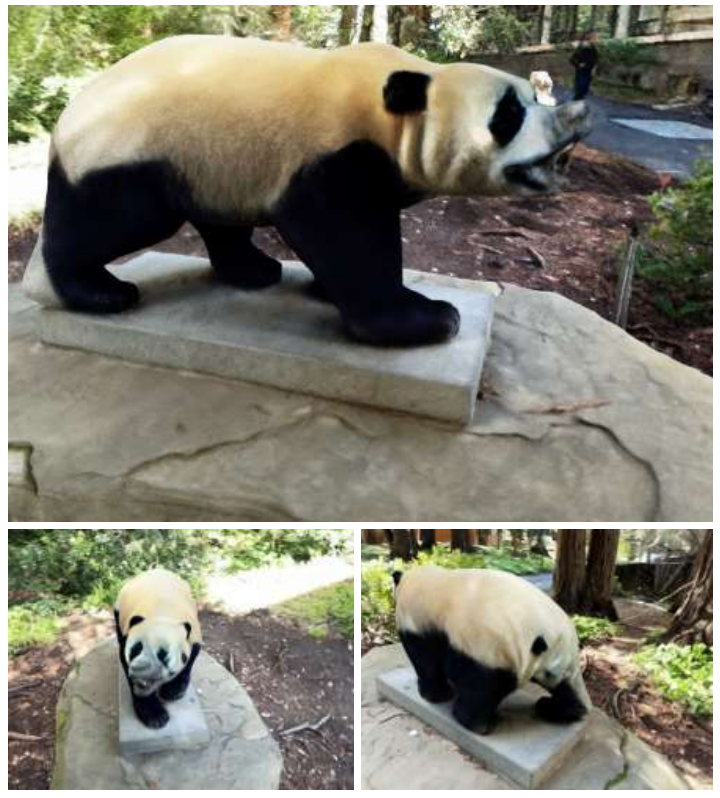}
        \end{minipage}
    \end{subfigure}

    \vspace{2px}

    \begin{subfigure}{0.24\linewidth}
        \begin{minipage}[t]{1.0\linewidth}
            \centering
            % \caption{Original}
            \includegraphics[width=1.0\linewidth]{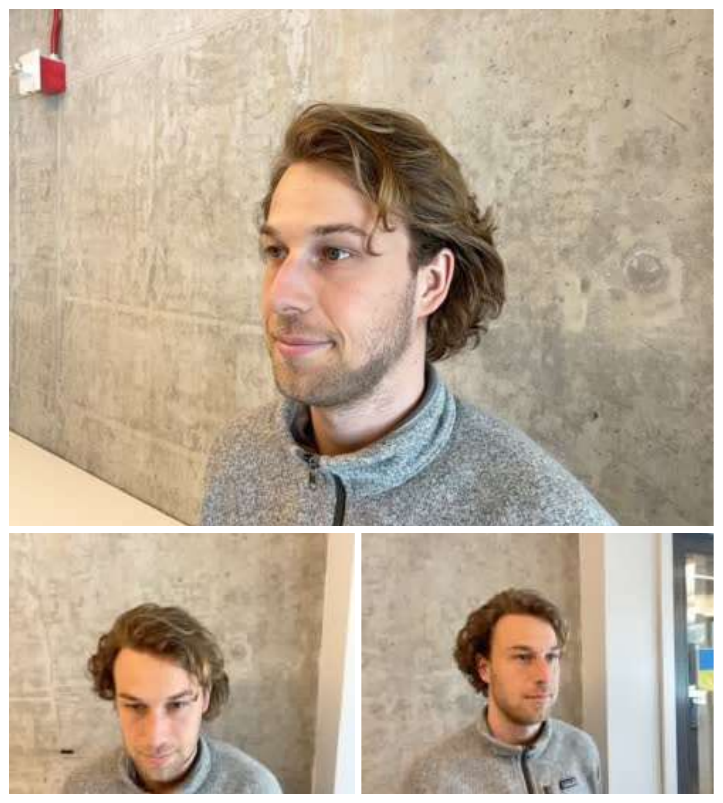}
        \end{minipage}
    \end{subfigure}
    \hfill
    \begin{subfigure}{0.24\linewidth}
        \begin{minipage}[t]{1.0\linewidth}
            \centering
            % \caption{IP2P}
            \includegraphics[width=1.0\linewidth]{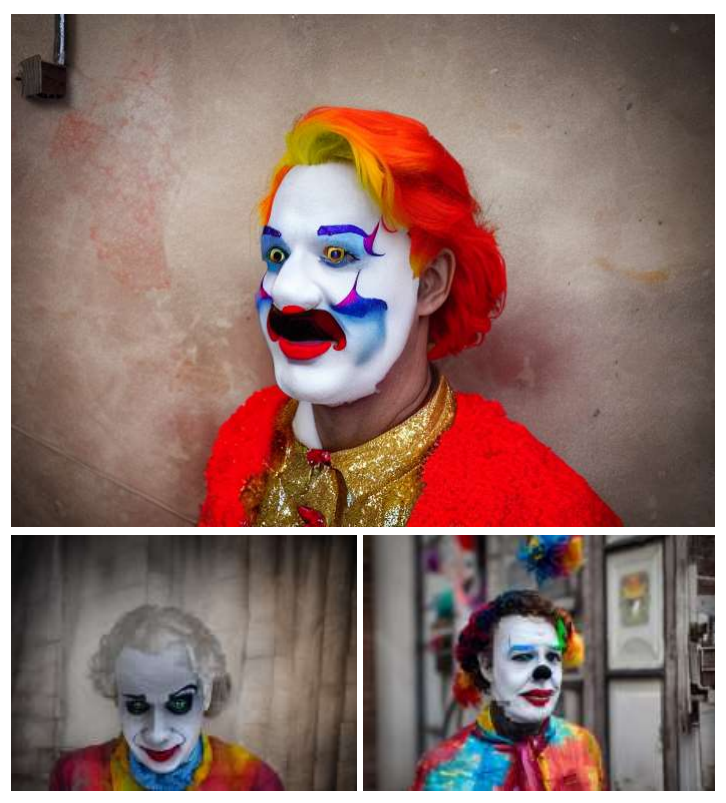}
        \end{minipage}
    \end{subfigure}
    \hfill
    \begin{subfigure}{0.24\linewidth}
        \begin{minipage}[t]{1.0\linewidth}
            \centering
            % \caption{ControlNet}
            \includegraphics[width=1.0\linewidth]{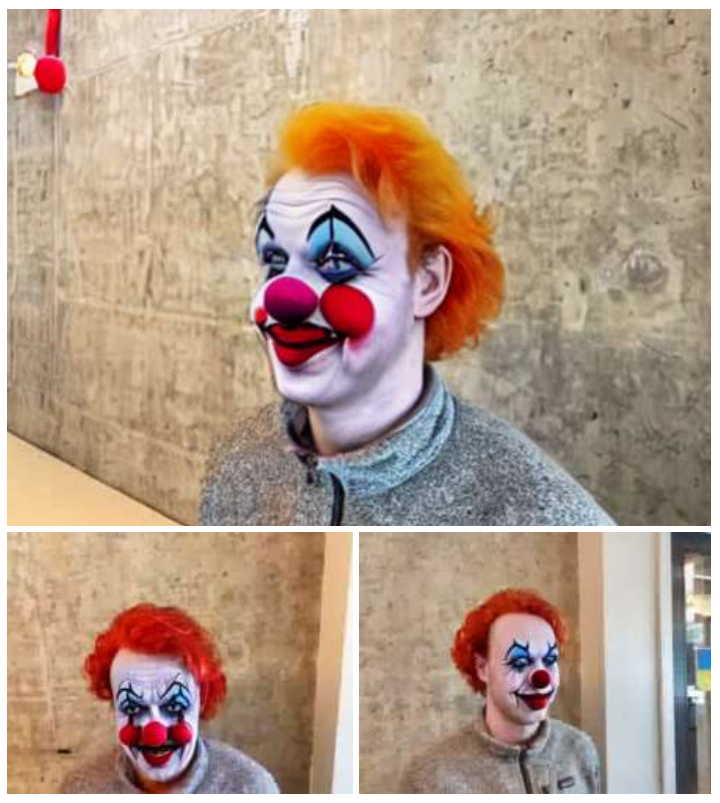}
        \end{minipage}
    \end{subfigure}
    \begin{subfigure}{0.24\linewidth}
        \begin{minipage}[t]{1.0\linewidth}
            \centering
            % \caption{DualNeRF}
            \includegraphics[width=1.0\linewidth]{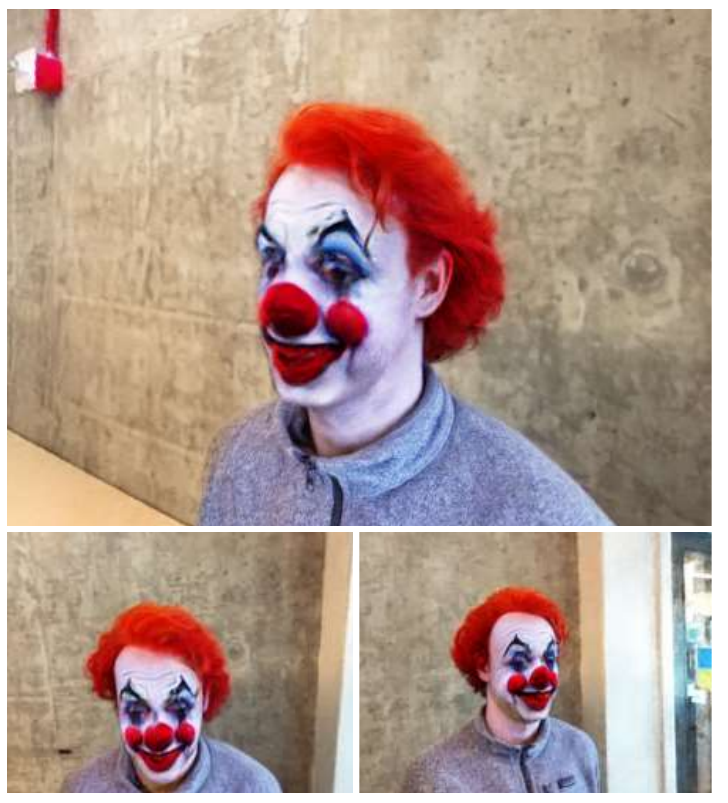}
        \end{minipage}
    \end{subfigure}
  
    % \caption{\textbf{Comparison with SOTA 2D Image Editing Methods.} The four columns respectively show the original scene and editing results from different views generated by ControlNet \cite{zhang2023adding}, IP2P \cite{brooks2023instructpix2pix}, and ours. The prompts used in two cases are "Turn the bear into a panda" and "Turn him into a clown" respectively. 2D methods generate inconsistent edits among different views. The editing quality of them is also unstable, resulting in low-quality edits and sometimes with wrong backgrounds.}
    \caption{\textbf{Comparison with SOTA 2D Image Editing Methods.} The four columns respectively show the original scene and editing results from different views generated by ControlNet \cite{zhang2023adding}, IP2P \cite{brooks2023instructpix2pix}, and ours. The prompts used in two cases are ``Turn the bear into a panda" and ``Turn him into a clown" respectively.}
    \label{fig: 2D}
\end{figure}

\subsection{Qualitative Results}
\paragraph{3D Scene Editing.}
The qualitative comparison between the editing results of DualNeRF with IN2N is shown in Fig. \ref{fig: qualitative results}. Both models train for $15k$ iterations for a fair comparison. Details in some results are zoomed in for a clearer observation. As illustrated in Fig. \ref{fig: qualitative results}, IN2N is hard to perform local edits as it cannot maintain non-target areas unaffected while editing the target areas. This results in blurry backgrounds, detail missing, and even artifacts.

Examples of blurry backgrounds can be seen in Fig. \ref{fig: 1-IN2N} and the first row of Fig. \ref{fig: qualitative results}, where blurred textures appear in the background areas of IN2N's edits. As a comparison, DualNeRF generates edits with clearer backgrounds, thanks to the additional guidance provided by our dual-field representation. Examples of detail missing are shown in the third to fifth rows in Fig. \ref{fig: qualitative results}, where details on the clothes, including the clothes texture, trousers pleats, and sweater weaving pattern, are faded away in edits of IN2N. These phenomena stem from IN2N's lack of efficient guidance to maintain details from original scenes. In contrast, DualNeRF finds a better balance between original image restoration and editing modification, preserving much more details than IN2N.
% Moreover, color drift often appears on unedited objects, such as the rock turning blue in the first row and the skin turning red in the fifth row in Fig. \ref{fig: qualitative results}. DualNeRF alleviates these issues successfully.
We also present two failure cases of IN2N in the second and last rows in Fig. \ref{fig: qualitative results}. Surprisingly, these outputs are different from the results displayed in \cite{haque2023instruct} under the default settings provided by their released code. However, DualNeRF generates much better edits under the same settings, demonstrating the superiority of our method.

\paragraph{Compared to 2D Methods.}
We compare DualNeRF with SOTA 2D image editing methods to demonstrate that pure 2D methods cannot edit 3D scenes with multi-view consistency. Fig. \ref{fig: 2D} demonstrates some examples of the comparison between the results of our edits with 2D methods. As we can see, ControlNet generates edits with low visual quality and high inconsistency among different views. Moreover, the background area is totally replaced by ControlNet, indicating that ControlNet cannot be used for local editing. IP2P edits the original scene with more background details preserved but still fails to generate edits with high multi-view consistency. As a comparison, consistent edits with restored backgrounds are generated by our method, thanks to the 3D nature of DualNeRF.

\begin{table}[t]
    \centering
    \rowcolors{1}{white}{gray!20}
    \begin{tabular}{p{70pt}<{\centering}|p{40pt}<{\centering}|p{40pt}<{\centering}|p{40pt}<{\centering}}
    \toprule
    Method & $C_{t2i} \uparrow$ & $C_{dir} \uparrow$ & SSIM $\uparrow$ \\
    \midrule
    per-frame IP2P & $0.2153$ & $0.9435$ & $\boldsymbol{0.8194}$ \\
    IN2N & $0.2170$ & $\boldsymbol{0.9806}$ & $0.7254$ \\
    DualNeRF & $\boldsymbol{0.2190}$ & $0.9777$ & $0.7362$ \\
    \bottomrule
    \end{tabular}
    % \caption{\textbf{Quantitative Evaluation.} We compare our method with baselines quantitatively based on CLIP. $C_{t2i}$ in the second column represents the CLIP text-image direction similarity \cite{haque2023instruct}, which measures the alignment between text prompt and scene edits. $C_{dir}$ in the third column denotes CLIP directional similarity \cite{haque2023instruct}, which indicates the consistency between adjacent views in the CLIP space. SSIM in the fourth column evaluates the edit's degree of restoration to the original image. Baseline methods include per-frame InstructPix2pix \cite{brooks2023instructpix2pix} and Instruct-NeRF2NeRF \cite{haque2023instruct}.}
    \caption{\textbf{Quantitative Evaluation.} We compare our method with baselines quantitatively based on CLIP. $C_{t2i}$ in the second column represents the CLIP text-image direction similarity \cite{haque2023instruct}. $C_{dir}$ in the third column denotes CLIP directional similarity \cite{haque2023instruct}. SSIM in the fourth column evaluates the edit's degree of restoration to the original image. Baseline methods include per-frame InstructPix2pix \cite{brooks2023instructpix2pix} and Instruct-NeRF2NeRF \cite{haque2023instruct}.}
    \label{tab: quantitative results}
\end{table}

\begin{figure*}
    \centering

    \begin{subfigure}{0.33\linewidth}
        \begin{minipage}[t]{1.0\linewidth}
            \centering
            \includegraphics[width=1.0\linewidth]{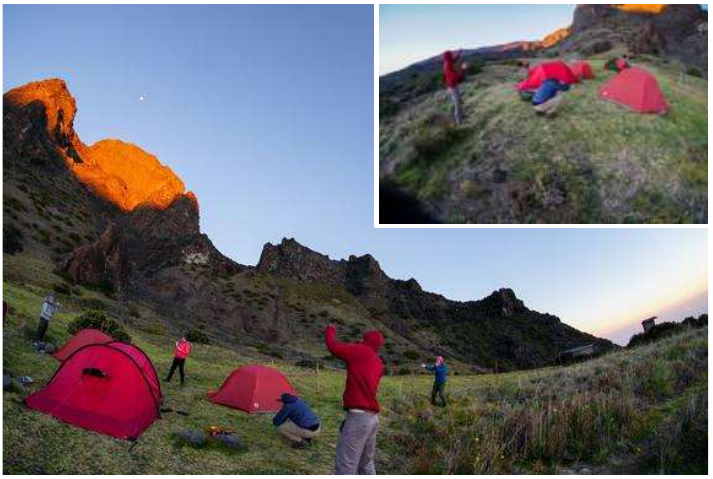}
            \caption{Original Scene}
        \end{minipage}
    \end{subfigure}
    \hfill
    \begin{subfigure}{0.33\linewidth}
        \begin{minipage}[t]{1.0\linewidth}
            \centering
            \includegraphics[width=1.0\linewidth]{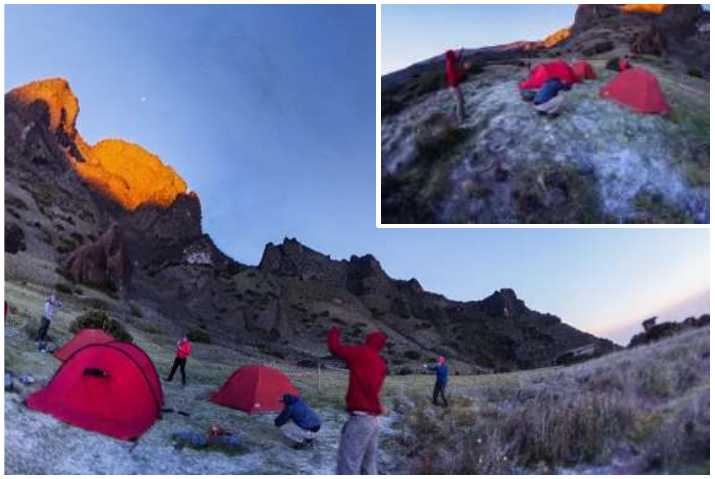}
            \caption{w/o SA and CCI}
        \end{minipage}
    \end{subfigure}
    \hfill
    \begin{subfigure}{0.33\linewidth}
        \begin{minipage}[t]{1.0\linewidth}
            \centering
            \includegraphics[width=1.0\linewidth]{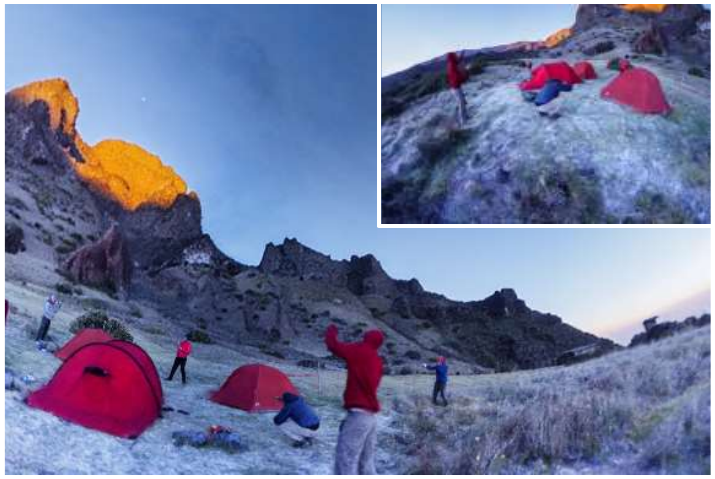}
            \caption{w/o CCI}
        \end{minipage}
    \end{subfigure}

    \begin{subfigure}{0.33\linewidth}
        \begin{minipage}[t]{1.0\linewidth}
            \centering
            \includegraphics[width=1.0\linewidth]{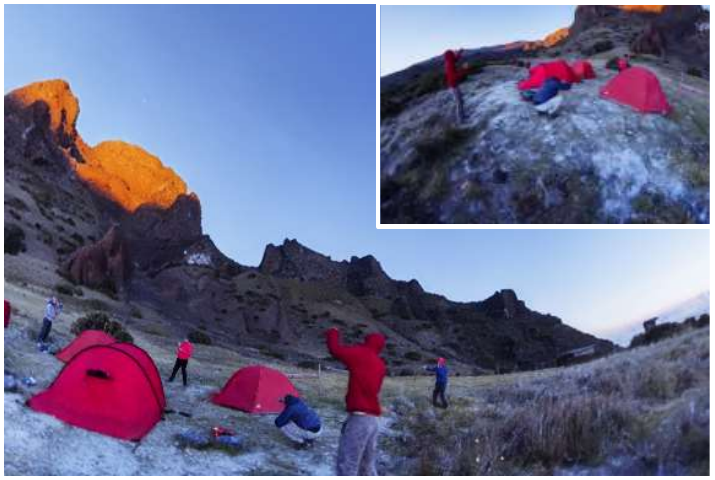}
            \caption{w/o DF and SA}
        \end{minipage}
    \end{subfigure}
    \hfill
    \begin{subfigure}{0.33\linewidth}
        \begin{minipage}[t]{1.0\linewidth}
            \centering
            \includegraphics[width=1.0\linewidth]{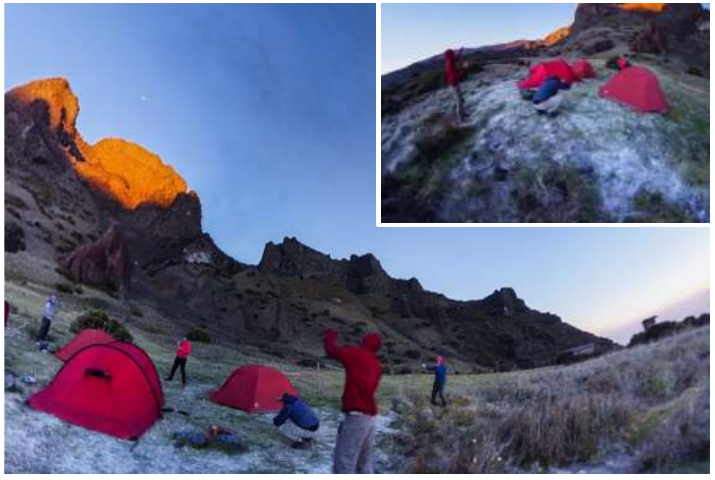}
            \caption{w/o SA}
        \end{minipage}
    \end{subfigure}
    \hfill
    \begin{subfigure}{0.33\linewidth}
        \begin{minipage}[t]{1.0\linewidth}
            \centering
            \includegraphics[width=1.0\linewidth]
            {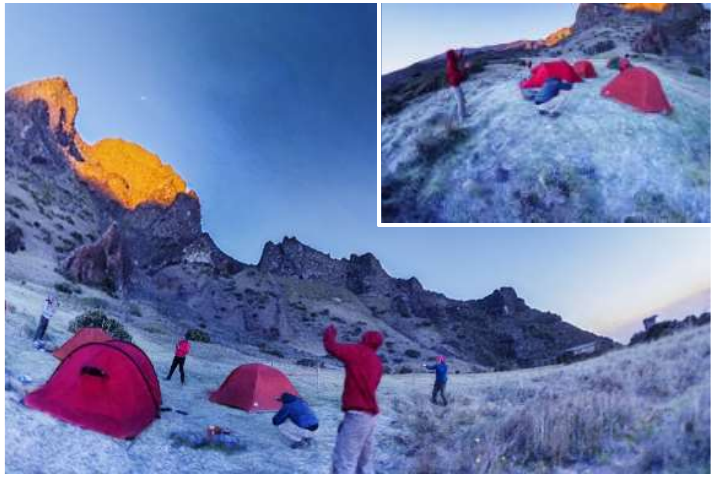}
            \caption{Full model}
        \end{minipage}
    \end{subfigure}
  
    \caption{\textbf{Ablation Study.} Qualitative results of our methods under different settings. Experiments are conducted in the \textit{campsite} scene conditioned on the prompt ``Make it look like it just snowed". DF, SA, and CCI represent dual-field representation, simulated annealing strategy, and CLIP-based consistency indicator respectively.}
    \label{fig: ablation study}
\end{figure*}

\subsection{Quantitative Results}
\label{sec: Quantitative Results}
Quantitative comparisons are also conducted between DualNeRF and baselines, as shown in Tab. \ref{tab: quantitative results}. Three metrics are used to evaluate the performance of edits, including a CLIP text-image direction similarity $C_{t2i}$, a CLIP directional similarity $C_{dir}$, and SSIM \cite{wang2004image}. Experiments are conducted across three scenes, including \textit{face}, \textit{fangzhou}, and \textit{person}, over $10$ edits. More details are provided in the supplementary material. As we can see in Tab. \ref{tab: quantitative results}, comparable performance are shown by DualNeRF and IN2N in the CLIP space. This shows that both methods generate edits with high text-to-image alignment ($C_{t2i}$) and multi-view consistency ($C_{dir}$). As a comparison, per-frame IP2P performs the worst in $C_{dir}$, indicating that 3D scenes are hard to edit solely by 2D methods. Additionally, DualNeRF outperforms IN2N in SSIM, which demonstrates that edits generated by our method present more restored backgrounds.

\begin{table}[t]
    \centering
    \rowcolors{1}{white}{gray!20}
    \begin{tabular}{p{30pt}<{\centering}|p{30pt}<{\centering}|p{30pt}<{\centering}|p{40pt}<{\centering}|p{40pt}<{\centering}}
    \toprule
    DF & SA & CCI & $C_{t2i} \uparrow$ & $C_{dir} \uparrow$ \\
    \midrule
    \multicolumn{3}{c|}{Original Scene} & $0.0244$ & $\boldsymbol{0.9383}$ \\
    \midrule
    \XSolidBrush & \XSolidBrush & \XSolidBrush & $0.1203$ & $0.9347$ \\
    \Checkmark & \XSolidBrush & \XSolidBrush & $0.1248$ & $0.9371$ \\
    \Checkmark & \Checkmark & \XSolidBrush & $0.1283$ & $0.9362$ \\
    \XSolidBrush & \XSolidBrush & \Checkmark & $0.1287$ & $0.9340$ \\
    \Checkmark & \XSolidBrush & \Checkmark & $0.1339$ & $0.9361$ \\
    \Checkmark & \Checkmark & \Checkmark & $\boldsymbol{0.1625}$ & $0.9352$ \\
    \bottomrule
    \end{tabular}
    \caption{\textbf{Ablation Study.} Experiments are conducted under different settings by removing some of our designs. DF, SA, and CCI represent dual-field representation, simulated annealing strategy, and CLIP-based consistency indicator respectively. Note that the row with three forks represents IN2N \cite{haque2023instruct}, while the row with three hooks represents our full model.}
    \label{tab: ablation study}
\end{table}

\subsection{Ablation Study}
The ablation study is also conducted to investigate the efficiency of our different designs. Specifically, we edit the \textit{campsite} scene conditioned on prompt ``Make it look like it just snowed" by DualNeRF under different settings. Qualitative results are shown in Fig. \ref{fig: ablation study}, while quantitative results can be seen in Tab. \ref{tab: ablation study}. As Fig. \ref{fig: ablation study} shows, models trained with our simulated annealing strategy successfully jump out of local optima ((c) and (f)), generating well-edited results compared to models without SA ((b), (d), and (e)). The use of the CLIP-based consistency indicator further improves the visual quality of the edits comparing examples in (c) and (f). The quantitative results in Tab. \ref{tab: ablation study} also confirm these points, as our full model's $C_{t2i}$ stands out as the best.

\section{Conclusion}
In this work, we propose DualNeRF, a novel text-driven 3D scene editing framework to perform local edits while preventing unwanted modification in irrelevant areas. Technically, we propose a dual-field architecture to provide additional guidance signal to the model during IDU, resulting in high-quality edits with restored backgrounds. Moreover, a simulated annealing strategy is introduced into the pipeline of IDU, helping the model address the local optima issue. A CLIP-based consistency indicator is also proposed to measure the edit consistency and filter out low-quality edits. Comprehensive experiments have demonstrated that our model outperforms previous works and displays strong power in 3D scene editing. We hope that DualNeRF can provide inspiration for subsequent work and pave the path to democratizing 3D content editing.

{
    \small
    \bibliographystyle{ieeenat_fullname}
    % \bibliography{main}

\begin{thebibliography}{57}
\providecommand{\natexlab}[1]{#1}
\providecommand{\url}[1]{\texttt{#1}}
\expandafter\ifx\csname urlstyle\endcsname\relax
  \providecommand{\doi}[1]{doi: #1}\else
  \providecommand{\doi}{doi: \begingroup \urlstyle{rm}\Url}\fi

\bibitem[Avrahami et~al.(2022)Avrahami, Lischinski, and Fried]{avrahami2022blended}
Omri Avrahami, Dani Lischinski, and Ohad Fried.
\newblock Blended diffusion for text-driven editing of natural images.
\newblock In \emph{Proceedings of the IEEE/CVF Conference on Computer Vision and Pattern Recognition}, pages 18208--18218, 2022.

\bibitem[Bar-Tal et~al.(2022)Bar-Tal, Ofri-Amar, Fridman, Kasten, and Dekel]{bar2022text2live}
Omer Bar-Tal, Dolev Ofri-Amar, Rafail Fridman, Yoni Kasten, and Tali Dekel.
\newblock Text2live: Text-driven layered image and video editing.
\newblock In \emph{European conference on computer vision}, pages 707--723. Springer, 2022.

\bibitem[Barron et~al.(2022)Barron, Mildenhall, Verbin, Srinivasan, and Hedman]{barron2022mip}
Jonathan~T Barron, Ben Mildenhall, Dor Verbin, Pratul~P Srinivasan, and Peter Hedman.
\newblock Mip-nerf 360: Unbounded anti-aliased neural radiance fields.
\newblock In \emph{Proceedings of the IEEE/CVF Conference on Computer Vision and Pattern Recognition}, pages 5470--5479, 2022.

\bibitem[Boss et~al.(2021)Boss, Jampani, Braun, Liu, Barron, and Lensch]{boss2021neural}
Mark Boss, Varun Jampani, Raphael Braun, Ce Liu, Jonathan Barron, and Hendrik Lensch.
\newblock Neural-pil: Neural pre-integrated lighting for reflectance decomposition.
\newblock \emph{Advances in Neural Information Processing Systems}, 34:\penalty0 10691--10704, 2021.

\bibitem[Brooks et~al.(2023)Brooks, Holynski, and Efros]{brooks2023instructpix2pix}
Tim Brooks, Aleksander Holynski, and Alexei~A Efros.
\newblock Instructpix2pix: Learning to follow image editing instructions.
\newblock In \emph{Proceedings of the IEEE/CVF Conference on Computer Vision and Pattern Recognition}, pages 18392--18402, 2023.

\bibitem[Brown et~al.(2020)Brown, Mann, Ryder, Subbiah, Kaplan, Dhariwal, Neelakantan, Shyam, Sastry, Askell, et~al.]{brown2020language}
Tom Brown, Benjamin Mann, Nick Ryder, Melanie Subbiah, Jared~D Kaplan, Prafulla Dhariwal, Arvind Neelakantan, Pranav Shyam, Girish Sastry, Amanda Askell, et~al.
\newblock Language models are few-shot learners.
\newblock \emph{Advances in neural information processing systems}, 33:\penalty0 1877--1901, 2020.

\bibitem[Couairon et~al.(2022)Couairon, Verbeek, Schwenk, and Cord]{couairon2022diffedit}
Guillaume Couairon, Jakob Verbeek, Holger Schwenk, and Matthieu Cord.
\newblock Diffedit: Diffusion-based semantic image editing with mask guidance.
\newblock \emph{arXiv preprint arXiv:2210.11427}, 2022.

\bibitem[Fang et~al.(2023)Fang, Wang, Yang, Tsai, Ding, Yang, and Zhou]{fang2023text}
Shuangkang Fang, Yufeng Wang, Yi Yang, Yi-Hsuan Tsai, Wenrui Ding, Ming-Hsuan Yang, and Shuchang Zhou.
\newblock Text-driven editing of 3d scenes without retraining.
\newblock \emph{arXiv preprint arXiv:2309.04917}, 2023.

\bibitem[Garbin et~al.(2022)Garbin, Kowalski, Estellers, Szymanowicz, Rezaeifar, Shen, Johnson, and Valentin]{garbin2022voltemorph}
Stephan~J Garbin, Marek Kowalski, Virginia Estellers, Stanislaw Szymanowicz, Shideh Rezaeifar, Jingjing Shen, Matthew Johnson, and Julien Valentin.
\newblock Voltemorph: Realtime, controllable and generalisable animation of volumetric representations.
\newblock \emph{arXiv preprint arXiv:2208.00949}, 2022.

\bibitem[Gong et~al.(2023)Gong, Wang, Han, and Dou]{gong2023recolornerf}
Bingchen Gong, Yuehao Wang, Xiaoguang Han, and Qi Dou.
\newblock Recolornerf: Layer decomposed radiance field for efficient color editing of 3d scenes.
\newblock \emph{arXiv preprint arXiv:2301.07958}, 2023.

\bibitem[Haque et~al.(2023)Haque, Tancik, Efros, Holynski, and Kanazawa]{haque2023instruct}
Ayaan Haque, Matthew Tancik, Alexei~A Efros, Aleksander Holynski, and Angjoo Kanazawa.
\newblock Instruct-nerf2nerf: Editing 3d scenes with instructions.
\newblock \emph{arXiv preprint arXiv:2303.12789}, 2023.

\bibitem[Hertz et~al.(2022)Hertz, Mokady, Tenenbaum, Aberman, Pritch, and Cohen-Or]{hertz2022prompt}
Amir Hertz, Ron Mokady, Jay Tenenbaum, Kfir Aberman, Yael Pritch, and Daniel Cohen-Or.
\newblock Prompt-to-prompt image editing with cross attention control.
\newblock \emph{arXiv preprint arXiv:2208.01626}, 2022.

\bibitem[Ho et~al.(2020)Ho, Jain, and Abbeel]{ho2020denoising}
Jonathan Ho, Ajay Jain, and Pieter Abbeel.
\newblock Denoising diffusion probabilistic models.
\newblock \emph{Advances in neural information processing systems}, 33:\penalty0 6840--6851, 2020.

\bibitem[Jambon et~al.(2023)Jambon, Kerbl, Kopanas, Diolatzis, Drettakis, and Leimk{\"u}hler]{jambon2023nerfshop}
Cl{\'e}ment Jambon, Bernhard Kerbl, Georgios Kopanas, Stavros Diolatzis, George Drettakis, and Thomas Leimk{\"u}hler.
\newblock Nerfshop: Interactive editing of neural radiance fields.
\newblock \emph{Proceedings of the ACM on Computer Graphics and Interactive Techniques}, 6\penalty0 (1), 2023.

\bibitem[Kawar et~al.(2023)Kawar, Zada, Lang, Tov, Chang, Dekel, Mosseri, and Irani]{kawar2023imagic}
Bahjat Kawar, Shiran Zada, Oran Lang, Omer Tov, Huiwen Chang, Tali Dekel, Inbar Mosseri, and Michal Irani.
\newblock Imagic: Text-based real image editing with diffusion models.
\newblock In \emph{Proceedings of the IEEE/CVF Conference on Computer Vision and Pattern Recognition}, pages 6007--6017, 2023.

\bibitem[Kirkpatrick et~al.(1983)Kirkpatrick, Gelatt~Jr, and Vecchi]{kirkpatrick1983optimization}
Scott Kirkpatrick, C~Daniel Gelatt~Jr, and Mario~P Vecchi.
\newblock Optimization by simulated annealing.
\newblock \emph{science}, 220\penalty0 (4598):\penalty0 671--680, 1983.

\bibitem[Kuang et~al.(2023)Kuang, Luan, Bi, Shu, Wetzstein, and Sunkavalli]{kuang2023palettenerf}
Zhengfei Kuang, Fujun Luan, Sai Bi, Zhixin Shu, Gordon Wetzstein, and Kalyan Sunkavalli.
\newblock Palettenerf: Palette-based appearance editing of neural radiance fields.
\newblock In \emph{Proceedings of the IEEE/CVF Conference on Computer Vision and Pattern Recognition}, pages 20691--20700, 2023.

\bibitem[Lee and Kim(2023)]{lee2023ice}
Jae-Hyeok Lee and Dae-Shik Kim.
\newblock Ice-nerf: Interactive color editing of nerfs via decomposition-aware weight optimization.
\newblock In \emph{Proceedings of the IEEE/CVF International Conference on Computer Vision}, pages 3491--3501, 2023.

\bibitem[Li et~al.(2022)Li, Lin, Forsyth, Huang, and Wang]{li2022climatenerf}
Yuan Li, Zhi-Hao Lin, David Forsyth, Jia-Bin Huang, and Shenlong Wang.
\newblock Climatenerf: Physically-based neural rendering for extreme climate synthesis.
\newblock \emph{arXiv e-prints}, pages arXiv--2211, 2022.

\bibitem[Liu et~al.(2020)Liu, Gu, Zaw~Lin, Chua, and Theobalt]{liu2020neural}
Lingjie Liu, Jiatao Gu, Kyaw Zaw~Lin, Tat-Seng Chua, and Christian Theobalt.
\newblock Neural sparse voxel fields.
\newblock \emph{Advances in Neural Information Processing Systems}, 33:\penalty0 15651--15663, 2020.

\bibitem[Liu et~al.(2021)Liu, Zhang, Zhang, Zhang, Zhu, and Russell]{liu2021editing}
Steven Liu, Xiuming Zhang, Zhoutong Zhang, Richard Zhang, Jun-Yan Zhu, and Bryan Russell.
\newblock Editing conditional radiance fields.
\newblock In \emph{Proceedings of the IEEE/CVF international conference on computer vision}, pages 5773--5783, 2021.

\bibitem[Meng et~al.(2021)Meng, He, Song, Song, Wu, Zhu, and Ermon]{meng2021sdedit}
Chenlin Meng, Yutong He, Yang Song, Jiaming Song, Jiajun Wu, Jun-Yan Zhu, and Stefano Ermon.
\newblock Sdedit: Guided image synthesis and editing with stochastic differential equations.
\newblock \emph{arXiv preprint arXiv:2108.01073}, 2021.

\bibitem[Mildenhall et~al.(2021)Mildenhall, Srinivasan, Tancik, Barron, Ramamoorthi, and Ng]{mildenhall2021nerf}
Ben Mildenhall, Pratul~P Srinivasan, Matthew Tancik, Jonathan~T Barron, Ravi Ramamoorthi, and Ren Ng.
\newblock Nerf: Representing scenes as neural radiance fields for view synthesis.
\newblock \emph{Communications of the ACM}, 65\penalty0 (1):\penalty0 99--106, 2021.

\bibitem[Mirzaei et~al.(2023)Mirzaei, Aumentado-Armstrong, Brubaker, Kelly, Levinshtein, Derpanis, and Gilitschenski]{mirzaei2023watch}
Ashkan Mirzaei, Tristan Aumentado-Armstrong, Marcus~A Brubaker, Jonathan Kelly, Alex Levinshtein, Konstantinos~G Derpanis, and Igor Gilitschenski.
\newblock Watch your steps: Local image and scene editing by text instructions.
\newblock \emph{arXiv preprint arXiv:2308.08947}, 2023.

\bibitem[Mokady et~al.(2023)Mokady, Hertz, Aberman, Pritch, and Cohen-Or]{mokady2023null}
Ron Mokady, Amir Hertz, Kfir Aberman, Yael Pritch, and Daniel Cohen-Or.
\newblock Null-text inversion for editing real images using guided diffusion models.
\newblock In \emph{Proceedings of the IEEE/CVF Conference on Computer Vision and Pattern Recognition}, pages 6038--6047, 2023.

\bibitem[M{\"u}ller et~al.(2022)M{\"u}ller, Evans, Schied, and Keller]{muller2022instant}
Thomas M{\"u}ller, Alex Evans, Christoph Schied, and Alexander Keller.
\newblock Instant neural graphics primitives with a multiresolution hash encoding.
\newblock \emph{ACM Transactions on Graphics (ToG)}, 41\penalty0 (4):\penalty0 1--15, 2022.

\bibitem[Nichol et~al.(2021)Nichol, Dhariwal, Ramesh, Shyam, Mishkin, McGrew, Sutskever, and Chen]{nichol2021glide}
Alex Nichol, Prafulla Dhariwal, Aditya Ramesh, Pranav Shyam, Pamela Mishkin, Bob McGrew, Ilya Sutskever, and Mark Chen.
\newblock Glide: Towards photorealistic image generation and editing with text-guided diffusion models.
\newblock \emph{arXiv preprint arXiv:2112.10741}, 2021.

\bibitem[Niemeyer and Geiger(2021)]{niemeyer2021giraffe}
Michael Niemeyer and Andreas Geiger.
\newblock Giraffe: Representing scenes as compositional generative neural feature fields.
\newblock In \emph{Proceedings of the IEEE/CVF Conference on Computer Vision and Pattern Recognition}, pages 11453--11464, 2021.

\bibitem[Parmar et~al.(2023)Parmar, Kumar~Singh, Zhang, Li, Lu, and Zhu]{parmar2023zero}
Gaurav Parmar, Krishna Kumar~Singh, Richard Zhang, Yijun Li, Jingwan Lu, and Jun-Yan Zhu.
\newblock Zero-shot image-to-image translation.
\newblock In \emph{ACM SIGGRAPH 2023 Conference Proceedings}, pages 1--11, 2023.

\bibitem[Peng et~al.(2022)Peng, Yan, Liu, Cheng, Guan, Pan, Zhai, and Yang]{peng2022cagenerf}
Yicong Peng, Yichao Yan, Shengqi Liu, Yuhao Cheng, Shanyan Guan, Bowen Pan, Guangtao Zhai, and Xiaokang Yang.
\newblock Cagenerf: Cage-based neural radiance field for generalized 3d deformation and animation.
\newblock \emph{Advances in Neural Information Processing Systems}, 35:\penalty0 31402--31415, 2022.

\bibitem[Radford et~al.(2021)Radford, Kim, Hallacy, Ramesh, Goh, Agarwal, Sastry, Askell, Mishkin, Clark, et~al.]{radford2021learning}
Alec Radford, Jong~Wook Kim, Chris Hallacy, Aditya Ramesh, Gabriel Goh, Sandhini Agarwal, Girish Sastry, Amanda Askell, Pamela Mishkin, Jack Clark, et~al.
\newblock Learning transferable visual models from natural language supervision.
\newblock In \emph{International conference on machine learning}, pages 8748--8763. PMLR, 2021.

\bibitem[Ramesh et~al.(2022)Ramesh, Dhariwal, Nichol, Chu, and Chen]{ramesh2022hierarchical}
Aditya Ramesh, Prafulla Dhariwal, Alex Nichol, Casey Chu, and Mark Chen.
\newblock Hierarchical text-conditional image generation with clip latents.
\newblock \emph{arXiv preprint arXiv:2204.06125}, 1\penalty0 (2):\penalty0 3, 2022.

\bibitem[Rombach et~al.(2022)Rombach, Blattmann, Lorenz, Esser, and Ommer]{rombach2022high}
Robin Rombach, Andreas Blattmann, Dominik Lorenz, Patrick Esser, and Bj{\"o}rn Ommer.
\newblock High-resolution image synthesis with latent diffusion models.
\newblock In \emph{Proceedings of the IEEE/CVF conference on computer vision and pattern recognition}, pages 10684--10695, 2022.

\bibitem[Ronneberger et~al.(2015)Ronneberger, Fischer, and Brox]{ronneberger2015u}
Olaf Ronneberger, Philipp Fischer, and Thomas Brox.
\newblock U-net: Convolutional networks for biomedical image segmentation.
\newblock In \emph{Medical Image Computing and Computer-Assisted Intervention--MICCAI 2015: 18th International Conference, Munich, Germany, October 5-9, 2015, Proceedings, Part III 18}, pages 234--241. Springer, 2015.

\bibitem[Saharia et~al.(2022)Saharia, Chan, Saxena, Li, Whang, Denton, Ghasemipour, Gontijo~Lopes, Karagol~Ayan, Salimans, et~al.]{saharia2022photorealistic}
Chitwan Saharia, William Chan, Saurabh Saxena, Lala Li, Jay Whang, Emily~L Denton, Kamyar Ghasemipour, Raphael Gontijo~Lopes, Burcu Karagol~Ayan, Tim Salimans, et~al.
\newblock Photorealistic text-to-image diffusion models with deep language understanding.
\newblock \emph{Advances in Neural Information Processing Systems}, 35:\penalty0 36479--36494, 2022.

\bibitem[Sch\"{o}nberger and Frahm(2016)]{schoenberger2016sfm}
Johannes~Lutz Sch\"{o}nberger and Jan-Michael Frahm.
\newblock Structure-from-motion revisited.
\newblock In \emph{Conference on Computer Vision and Pattern Recognition (CVPR)}, 2016.

\bibitem[Sella et~al.(2023)Sella, Fiebelman, Hedman, and Averbuch-Elor]{sella2023vox}
Etai Sella, Gal Fiebelman, Peter Hedman, and Hadar Averbuch-Elor.
\newblock Vox-e: Text-guided voxel editing of 3d objects.
\newblock In \emph{Proceedings of the IEEE/CVF International Conference on Computer Vision}, pages 430--440, 2023.

\bibitem[Sohl-Dickstein et~al.(2015)Sohl-Dickstein, Weiss, Maheswaranathan, and Ganguli]{sohl2015deep}
Jascha Sohl-Dickstein, Eric Weiss, Niru Maheswaranathan, and Surya Ganguli.
\newblock Deep unsupervised learning using nonequilibrium thermodynamics.
\newblock In \emph{International conference on machine learning}, pages 2256--2265. PMLR, 2015.

\bibitem[Song et~al.(2020)Song, Meng, and Ermon]{song2020denoising}
Jiaming Song, Chenlin Meng, and Stefano Ermon.
\newblock Denoising diffusion implicit models.
\newblock \emph{arXiv preprint arXiv:2010.02502}, 2020.

\bibitem[Song et~al.(2022)Song, Han, Liu, Metaxas, and Elgammal]{song2022diffusion}
Kunpeng Song, Ligong Han, Bingchen Liu, Dimitris Metaxas, and Ahmed Elgammal.
\newblock Diffusion guided domain adaptation of image generators.
\newblock \emph{arXiv preprint arXiv:2212.04473}, 2022.

\bibitem[Srinivasan et~al.(2021)Srinivasan, Deng, Zhang, Tancik, Mildenhall, and Barron]{srinivasan2021nerv}
Pratul~P Srinivasan, Boyang Deng, Xiuming Zhang, Matthew Tancik, Ben Mildenhall, and Jonathan~T Barron.
\newblock Nerv: Neural reflectance and visibility fields for relighting and view synthesis.
\newblock In \emph{Proceedings of the IEEE/CVF Conference on Computer Vision and Pattern Recognition}, pages 7495--7504, 2021.

\bibitem[Tancik et~al.(2023)Tancik, Weber, Ng, Li, Yi, Kerr, Wang, Kristoffersen, Austin, Salahi, Ahuja, McAllister, and Kanazawa]{nerfstudio}
Matthew Tancik, Ethan Weber, Evonne Ng, Ruilong Li, Brent Yi, Justin Kerr, Terrance Wang, Alexander Kristoffersen, Jake Austin, Kamyar Salahi, Abhik Ahuja, David McAllister, and Angjoo Kanazawa.
\newblock Nerfstudio: A modular framework for neural radiance field development.
\newblock In \emph{ACM SIGGRAPH 2023 Conference Proceedings}, 2023.

\bibitem[Tertikas et~al.(2023)Tertikas, Despoina, Pan, Park, Uy, Emiris, Avrithis, and Guibas]{tertikas2023partnerf}
Konstantinos Tertikas, Pascalidou Despoina, Boxiao Pan, Jeong~Joon Park, Mikaela~Angelina Uy, Ioannis Emiris, Yannis Avrithis, and Leonidas Guibas.
\newblock Partnerf: Generating part-aware editable 3d shapes without 3d supervision.
\newblock \emph{arXiv preprint arXiv:2303.09554}, 2023.

\bibitem[Valevski et~al.(2022)Valevski, Kalman, Matias, and Leviathan]{valevski2022unitune}
Dani Valevski, Matan Kalman, Yossi Matias, and Yaniv Leviathan.
\newblock Unitune: Text-driven image editing by fine tuning an image generation model on a single image.
\newblock \emph{arXiv preprint arXiv:2210.09477}, 2022.

\bibitem[Wang et~al.(2022)Wang, Chai, He, Chen, and Liao]{wang2022clip}
Can Wang, Menglei Chai, Mingming He, Dongdong Chen, and Jing Liao.
\newblock Clip-nerf: Text-and-image driven manipulation of neural radiance fields.
\newblock In \emph{Proceedings of the IEEE/CVF Conference on Computer Vision and Pattern Recognition}, pages 3835--3844, 2022.

\bibitem[Wang et~al.(2023)Wang, Jiang, Chai, He, Chen, and Liao]{wang2023nerf}
Can Wang, Ruixiang Jiang, Menglei Chai, Mingming He, Dongdong Chen, and Jing Liao.
\newblock Nerf-art: Text-driven neural radiance fields stylization.
\newblock \emph{IEEE Transactions on Visualization and Computer Graphics}, 2023.

\bibitem[Wang et~al.(2004)Wang, Bovik, Sheikh, and Simoncelli]{wang2004image}
Zhou Wang, Alan~C Bovik, Hamid~R Sheikh, and Eero~P Simoncelli.
\newblock Image quality assessment: from error visibility to structural similarity.
\newblock \emph{IEEE transactions on image processing}, 13\penalty0 (4):\penalty0 600--612, 2004.

\bibitem[Xiang et~al.(2021)Xiang, Xu, Hasan, Hold-Geoffroy, Sunkavalli, and Su]{xiang2021neutex}
Fanbo Xiang, Zexiang Xu, Milos Hasan, Yannick Hold-Geoffroy, Kalyan Sunkavalli, and Hao Su.
\newblock Neutex: Neural texture mapping for volumetric neural rendering.
\newblock In \emph{Proceedings of the IEEE/CVF Conference on Computer Vision and Pattern Recognition}, pages 7119--7128, 2021.

\bibitem[Xu and Harada(2022)]{xu2022deforming}
Tianhan Xu and Tatsuya Harada.
\newblock Deforming radiance fields with cages.
\newblock In \emph{European Conference on Computer Vision}, pages 159--175. Springer, 2022.

\bibitem[Xu et~al.(2023)Xu, Chai, Shi, Peng, Skorokhodov, Siarohin, Yang, Shen, Lee, Zhou, et~al.]{xu2023discoscene}
Yinghao Xu, Menglei Chai, Zifan Shi, Sida Peng, Ivan Skorokhodov, Aliaksandr Siarohin, Ceyuan Yang, Yujun Shen, Hsin-Ying Lee, Bolei Zhou, et~al.
\newblock Discoscene: Spatially disentangled generative radiance fields for controllable 3d-aware scene synthesis.
\newblock In \emph{Proceedings of the IEEE/CVF Conference on Computer Vision and Pattern Recognition}, pages 4402--4412, 2023.

\bibitem[Yuan et~al.(2022)Yuan, Sun, Lai, Ma, Jia, and Gao]{yuan2022nerf}
Yu-Jie Yuan, Yang-Tian Sun, Yu-Kun Lai, Yuewen Ma, Rongfei Jia, and Lin Gao.
\newblock Nerf-editing: geometry editing of neural radiance fields.
\newblock In \emph{Proceedings of the IEEE/CVF Conference on Computer Vision and Pattern Recognition}, pages 18353--18364, 2022.

\bibitem[Zhang et~al.(2020)Zhang, Riegler, Snavely, and Koltun]{zhang2020nerf++}
Kai Zhang, Gernot Riegler, Noah Snavely, and Vladlen Koltun.
\newblock Nerf++: Analyzing and improving neural radiance fields.
\newblock \emph{arXiv preprint arXiv:2010.07492}, 2020.

\bibitem[Zhang et~al.(2022{\natexlab{a}})Zhang, Kolkin, Bi, Luan, Xu, Shechtman, and Snavely]{zhang2022arf}
Kai Zhang, Nick Kolkin, Sai Bi, Fujun Luan, Zexiang Xu, Eli Shechtman, and Noah Snavely.
\newblock Arf: Artistic radiance fields.
\newblock In \emph{European Conference on Computer Vision}, pages 717--733. Springer, 2022{\natexlab{a}}.

\bibitem[Zhang et~al.(2023)Zhang, Rao, and Agrawala]{zhang2023adding}
Lvmin Zhang, Anyi Rao, and Maneesh Agrawala.
\newblock Adding conditional control to text-to-image diffusion models, 2023.

\bibitem[Zhang et~al.(2018)Zhang, Isola, Efros, Shechtman, and Wang]{8578166}
Richard Zhang, Phillip Isola, Alexei~A. Efros, Eli Shechtman, and Oliver Wang.
\newblock The unreasonable effectiveness of deep features as a perceptual metric.
\newblock In \emph{2018 IEEE/CVF Conference on Computer Vision and Pattern Recognition}, pages 586--595, 2018.

\bibitem[Zhang et~al.(2021)Zhang, Srinivasan, Deng, Debevec, Freeman, and Barron]{zhang2021nerfactor}
Xiuming Zhang, Pratul~P Srinivasan, Boyang Deng, Paul Debevec, William~T Freeman, and Jonathan~T Barron.
\newblock Nerfactor: Neural factorization of shape and reflectance under an unknown illumination.
\newblock \emph{ACM Transactions on Graphics (ToG)}, 40\penalty0 (6):\penalty0 1--18, 2021.

\bibitem[Zhang et~al.(2022{\natexlab{b}})Zhang, Han, Ghosh, Metaxas, and Ren]{zhang2022sine}
Zhixing Zhang, Ligong Han, Arnab Ghosh, Dimitris Metaxas, and Jian Ren.
\newblock Sine: Single image editing with text-to-image diffusion models.
\newblock \emph{arXiv preprint arXiv:2212.04489}, 2022{\natexlab{b}}.

\end{thebibliography}

}

% WARNING: do not forget to delete the supplementary pages from your submission 
% \input{sec/X_suppl}

% \input{sec/X_suppl.tex}

\end{document}